\documentclass[10pt,conference,letterpaper]{IEEEtran}
\IEEEoverridecommandlockouts
\usepackage{algorithm}
\usepackage{multirow}
\usepackage{amsmath, amssymb, amsfonts}
\usepackage{amsthm}
\usepackage{graphicx}
\usepackage{setspace}
\usepackage{multicol}
\usepackage{multirow}
\usepackage{tabularx}
\usepackage{makecell}
\usepackage{cite}
\usepackage{float}
\usepackage{alltt, listings}
\usepackage{subfigure}
\usepackage{algpseudocode}
\usepackage{wrapfig}
\usepackage{times}
\usepackage{fancyhdr}
\usepackage{enumitem}
\usepackage{url}
\usepackage{filecontents}
\usepackage{bbding}
\usepackage{color}

\newtheorem{assumption}{Assumption}
\newtheorem{theorem}{Theorem}
\newtheorem{lemma}{Lemma}

\newtheorem{proposition}{Proposition}
\newtheorem{remark}{Remark}
\theoremstyle{definition}
\newtheorem{definition}{Definition}

\begin {document}

\title{Byzantine-robust Federated Learning through Collaborative Malicious Gradient Filtering}

\author{
%	Anonymous Author(s)
{\rm Jian Xu$^1$, Shao-Lun Huang$^{1*}$\thanks{$^*$ Corresponding author: shaolun.huang@sz.tsinghua.edu.cn}, Linqi Song$^2$, Tian Lan$^3$} \\
$^1$Tsinghua University, $^2$City University of Hong Kong, $^3$George Washington University \\
}

\maketitle

\begin{abstract}
	Gradient-based training in federated learning is known to be vulnerable to faulty/malicious clients, which are often modeled as Byzantine clients. To this end, previous work either makes use of auxiliary data at parameter server to verify the received gradients (e.g., by computing validation error rate) or leverages statistic-based methods (e.g. median and Krum) to identify and remove malicious gradients from Byzantine clients. In this paper, we remark that auxiliary data may not always be available in practice and focus on the statistic-based approach. However, recent work on model poisoning attacks has shown that well-crafted attacks can circumvent most of median- and distance-based statistical defense methods, making malicious gradients indistinguishable from honest ones. To tackle this challenge, we show that the element-wise sign of gradient vector can provide valuable insight in detecting model poisoning attacks. Based on our theoretical analysis of the \textit{Little is Enough} attack, we propose a novel approach called \textit{SignGuard} to enable Byzantine-robust federated learning through collaborative malicious gradient filtering. More precisely, the received gradients are first processed to generate relevant magnitude, sign, and similarity statistics, which are then collaboratively utilized by multiple filters to eliminate malicious gradients before final aggregation. Finally, extensive experiments of image and text classification tasks are conducted under recently proposed attacks and defense strategies. The numerical results demonstrate the effectiveness and superiority of our proposed approach. The code is available at \textit{\url{https://github.com/JianXu95/SignGuard}}

\end{abstract}

\section{Introduction}
\label{sec:intro}

\begin{figure}[t]	
	{	
		\begin{center}
			%\hspace{1ex}
			\includegraphics[width=0.9\columnwidth,clip=true]{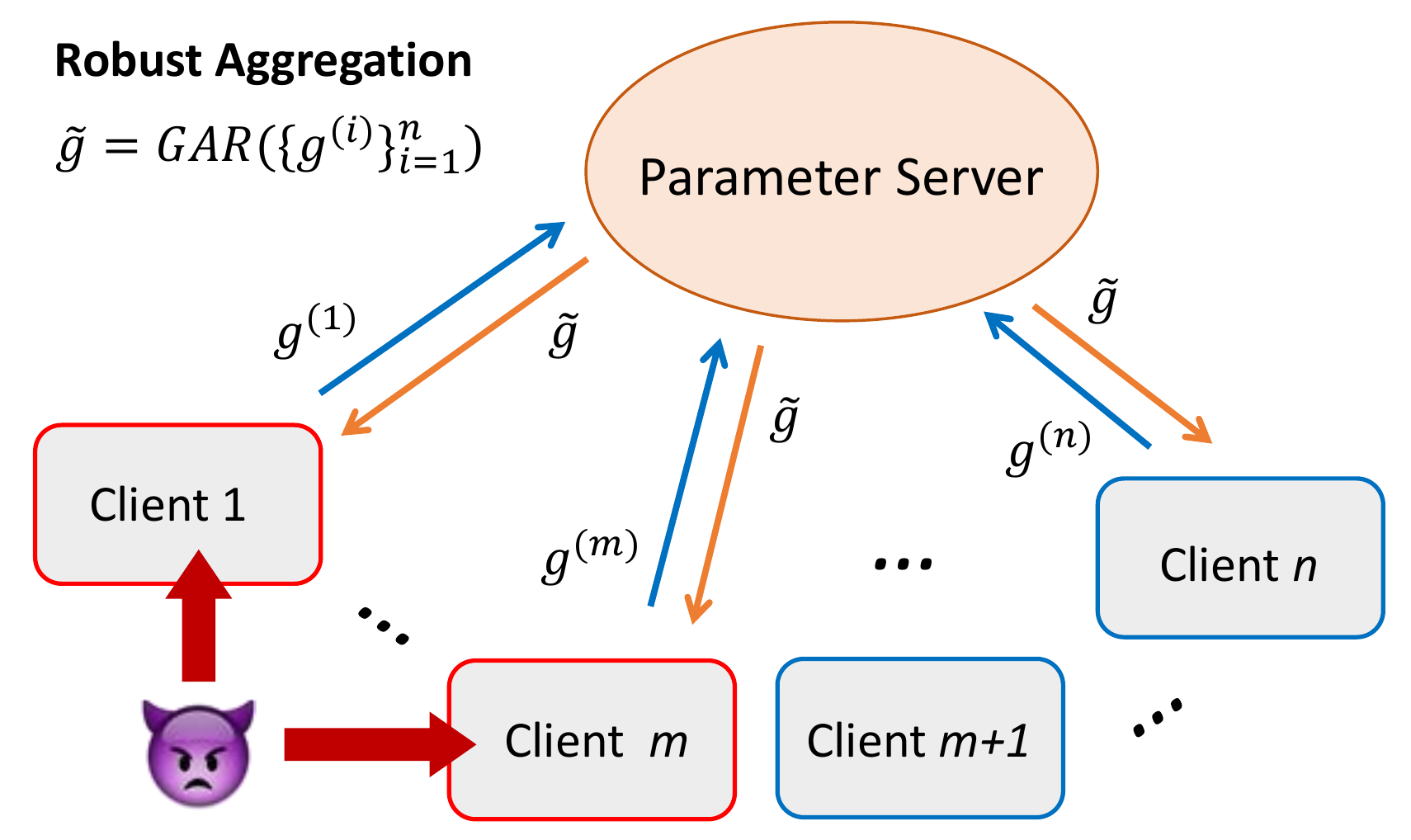}
		\end{center}
	}
	%	\vspace{-3ex}
	\caption{Federated learning system: one parameter server with $n$ clients, in which a attacker controls $m$ Byzantine clients to attack the learning system.}
		\vspace{-2ex}
	\label{fig:example}
\end{figure}

In the era of big data, private data are often scattered among local clients (e.g., companies, mobile devices), leading to the problem of isolated data islands \cite{yang2019federated}. To fully capitalize on the value of big data while protecting data privacy and security, federated learning (FL) has attracted significant interest in recent years \cite{yang2019FLbook, mcmahan17,yang2019federated,Kairouz21AdvanceFL,Li20FL}. A typical setup of FL consists of a parameter server (PS) and a number of distributed clients, where the local training data are prohibited from sharing among the clients. The general goal of FL is to jointly train a global model that has high generalization ability than that only trained on local data. While FL systems allow clients to keep their private data local, a significant vulnerability arises when a subset of clients aim to prevent successful training of the global model, which are modeled as Byzantine clients \cite{Blanchard17Byz,lyu20survey,lyu20privacy_robust}. This can be seen through a simple example shown in Fig.~\ref{fig:example} with one PS and $n-m$ benign clients as well as $m$ Byzantine clients, where the Byzantine clients can send arbitrary model update vectors to the PS, which may significantly poison the training process if not identified and removed by PS.  It has been shown that mitigating Byzantine model poisoning attacks is crucial for robust FL and other distributed learning \cite{shenTS16auror,Blanchard17Byz,Fang20Local}. On the other hand, distributed implementation of gradient-based learning algorithms \cite{lecun1998gd} are increasingly popular for training large-scale models on distributed datasets, e.g., deep neural networks for human face identification and news sentimental analysis \cite{krizhevsky2012imagenet,hinton2012speech,dean2012large}. Therefore, many efforts have been devoted to developing robust \textbf{\underline{g}}radient \textbf{\underline{a}}ggregation \textbf{\underline{r}}ules (GAR) \cite{Kairouz21AdvanceFL} to achieve Byzantine-robust FL algorithms.

Recently, much research attention has focused on mitigating Byzantine attacks either by leveraging statistic-based outlier detection techniques \cite{Chen17BGD,Yin18optimalrate} or by utilizing auxiliary labeled data collected by PS to verify the correctness of received gradients \cite{Xie19Zeno,CaoL19arbitraryByz}. We remark that auxiliary data sufficiently capturing the global data distribution may not be practicable to PS. And recent works have shown that existing statistic-based aggregation rules are vulnerable to well-crafted model poisoning attacks \cite{BaruchBG19LIE,shejwalkar2021manipulating}, which are indistinguishable in Euclidean distance such that they can circumvent most defenses.

In this paper, we focus on the gradient-based FL systems and propose a novel robust gradient aggregation framework, namely SignGuard, to enable Byzantine-robust federated learning. SignGuard leverages a new technique of sign-gradient filtering to identify malicious gradients and can be integrated with existing gradient aggregation rules, such as trimmed-mean \cite{Yin18optimalrate}. In particular, we define \emph{sign-gradient} as the element-wise sign of a gradient vector. The key idea of SignGuard is that the sign distribution of {sign-gradient} can provide valuable information in detecting advanced model poisoning attacks, which would otherwise evade state-of-the-art statistic-based detection methods such as Krum and Bulyan \cite{Blanchard17Byz,Mhamdi18Bulyan}. SignGuard is inspired by our theoretical analysis of \emph{Little is Enough} (LIE) attack \cite{BaruchBG19LIE}, and the generally good performance of signSGD \cite{bernstein18sign} in distributed learning tasks. In \cite{bernstein18sign} the authors show that even if PS only collects the sign of gradient, the model training can still converge with small accuracy degradation and keep the training process fault-tolerant. This fact tells us that the sign of gradient plays a vital role in model updating. Our novel analysis on the LIE attack reveals that gradient manipulation can cause {significant variation of sign distribution}, which turns out to be a breakthrough against such well-crafted attacks. We also empirically find that even the simplest \textit{sign statistics}\footnote{By default, we use the ``sign statistics'' to denote the proportions of positive, negative, and zero signs.} can expose most of the attacks. These observations provide a new perspective towards Byzantine attack mitigation and directly inspire the design of our SignGuard framework. The core of our approach is extracting robust features of received gradients and using an unsupervised clustering method to remove the anomalous ones. We find this simple and practical strategy can detect suspicious gradients effectively and efficiently. 

To the best of our knowledge, this is the first work to utilize sign statistics of gradients for Byzantine-robust federated learning. SignGuard employs well-designed filtering techniques to identify and eliminate the suspicious gradients to favor gradient aggregation. Our theoretical analysis proves that SignGuard can guarantee training convergence on both IID and non-IID training data while introducing no extra overhead for local computation or auxiliary data collection. In particular, for a system with $n$ clients including $m$ Byzantine clients and satisfying $n\ge 2m+1$, we quantify the gradient bias induced by ignoring $m$ suspicious gradients and show that the parameters enjoy a similar update rule as in safe training, thus the convergence analysis could be performed similarly. Finally, the SignGuard framework is evaluated on various real-world image and text classification tasks through extensive experiments by changing the attacks and the percentages of malicious clients. Evaluation results demonstrate the effectiveness of our SignGuard in protecting the FL system from Byzantine poisoning attacks and meanwhile achieving high model accuracy. To summarize, we make the following key contributions: 
\begin{itemize}[leftmargin=3ex]
\item A novel gradient aggregation framework called SignGuard is proposed for Byzantine-robust federated learning, which leverages the sign statistics of gradients to defend against model poisoning attacks.

\item We provide a theoretical analysis of the harmfulness and stealthiness of the state-of-the-art \textit{Little is Enough} attack and also propose a new hybrid attack strategy.

\item The convergence of SignGuard is proven with a appropriate choice of learning rate. In particular, we show that Byzantine clients inevitably affect the convergence error in non-IID settings even if all malicious gradients are removed.

\item SignGuard is verified through extensive experiments on MNIST/Fashion-MNIST, CIFAR-10, and AG-News datasets under various Byzantine attacks. Compared with existing approaches, our method exhibits superior performance in both IID and non-IID settings.
\end{itemize}

\section{Background and Related Work}
\label{secRelated} 

\subsection{Safety \& Security in Federated Learning}

The model safety and data security are essential principles of federated learning due to the concern of privacy risks and adversarial threats \cite{yang2019federated, Kairouz21AdvanceFL, lyu20survey, ma20safeFL}, especially under tough privacy regulations such as General Data
Protection Regulation (GDPR) \cite{sharma2019data}. Meanwhile, the learning systems are vulnerable to various kinds of failures, including non-malicious faults and malicious attacks. Data poisoning and model update poisoning attacks aim to degrade or even fully break the global model during the training phase, while backdoor attacks (aka. targeted attacks) make the model misclassify certain samples during the inference phase \cite{Kairouz21AdvanceFL}.
In particular, the Byzantine threats can be viewed as worst-case attacks, in which corrupted clients can produce arbitrary outputs and are allowed to collude \cite{Blanchard17Byz, Xie19Empires, shejwalkar2021manipulating}.

\subsection{Existing Defense Strategies}
Existing defenses either leverage statistic-based robust aggregation rule to get reliable gradient estimation, or utilize auxiliary data in PS to validate the received gradients. The former is also known as majority-vote based strategy and requiring the percentage of Byzantine clients less than 50\%, such as {Krum}  \cite{Blanchard17Byz}, {trimmed-mean (TrMean)} and {coordinate-wise median (Median)} \cite{Yin18optimalrate} and {Bulyan} \cite{Mhamdi18Bulyan}. Specially, some works only aggregate the sign of gradient to mitigate the Byzantine effect \cite{bernstein18sign, li2019rsa}. Recently, a method called {Divider and Conquer} (DnC) is proposed to tackle strong attacks \cite{shejwalkar2021manipulating}. When auxiliary data is available in PS, robustness can be guaranteed by validating the performance of received gradients/models. {Zeno} \cite{Xie19Zeno} use a stochastic descendant score to evaluate the correctness of each gradient and choose those with the highest scores. Fang \cite{Fang20Local} use error rate based and loss function based rejection mechanism to reject gradients that have a bad impact on model updating. In \cite{cao20FLTrust}, the authors utilize the ReLU-clipped cosine-similarity between each received gradient and standard gradient as the weight to get robust aggregation. The main concern of such approaches is the accessibility of auxiliary data and the extra computational overhead.

Some studies show that malicious behavior could be revealed from the gradient trace by designing advanced filter techniques \cite{Alistarh18Byz, zhu20safeguard, Mu19AFA}. Besides, the client-side momentum SGD can also be considered as a history-aided method and can help to alleviate the impact of Byzantine attacks\cite{Karimireddy20history,Mahdi21momentum}. Another line of work utilizes data redundancy to eliminate the effect of Byzantine failures. In \cite{Chen18Draco}, the authors present a scalable framework called {DRACO} for robust distributed training using ideas from coding theory. In \cite{Rajput19Detox}, a framework called {DETOX} is proposed by combing computational redundancy and hierarchical robust aggregation to filter out Byzantine gradients. In \cite{SohnHCM20codedsignsgd}, signSGD with election coding is proposed for robust and communication-efficient distributed learning. Moreover, provable security guarantee is also explored in \cite{SteinhardtKL17, CaoJG21}.

\section{Rethinking of LIE Attack}
\label{secAttackAnalysis}

In this section, we present our theoretical analysis along with empirical evidence of the \emph{Little is Enough} (LIE) attack \cite{BaruchBG19LIE} to demonstrate the limitation of existing median- and distance-based defenses.

\vspace{1ex}
\noindent \textbf{{LIE} Attack.} Byzantine clients first estimate coordinate-wise mean ($\mu_j$) and standard deviation ($\sigma_j$), and then send malicious gradient vector with elements crafted as follows:
\begin{equation}\label{eq:lie}
(g_m)_j = \mu_j - z\cdot\sigma_j, ~ j \in [d]
\end{equation}
where the positive attack factor $z$ depends on the total number of clients and Byzantine fraction, and can be determined by using cumulative standard normal function $\phi(z)$:
\begin{equation}
z_{max} =  max_z \left(\phi(z)<\frac{n-\left\lfloor\frac{n}{2}+1\right\rfloor}{n-m}\right)
\end{equation}

In the following, we will show why this attack is harmful and hard to detect. Recall that signSGD can achieve good model accuracy by only utilizing the sign of gradient, which illuminates a fact that the sign of gradient plays a crucial role in model updating. Therefore, it's important to check the sign of gradient for this type of attack. The crafting rule of {LIE} attack is already shown in Eq. (\ref{eq:lie}), from which we can see that $(g_m)_j $ could have opposite sign with $\mu_j$ when $\mu_j>0$. For coordinate-wise median and $\mu_j>0$, we assume this aggregation rule results in $\tilde{g} = g_m$, then we have:
\begin{equation}
if ~~ z > \frac{\mu_j}{\sigma_j}, ~~then ~~sign(\tilde{g}_j) \ne sign(\mu_j)
\end{equation} For mean aggregation rule and $\mu_j>0$, if $\mu_j$ and $\sigma_j$ are estimated on benign clients, then the $j$-th element becomes:
\begin{equation}
\tilde{g}_j = \frac{1}{n}[m\cdot (g_m)_j + (n-m)\mu_j] = \mu_j - z\cdot \beta \cdot\sigma_j
\end{equation}
and in this case a bigger $z$ is needed to reverse the sign:
\begin{equation}
if ~~ z > \frac{n\mu_j}{m\sigma_j}, ~~then ~~sign(\tilde{g}_j) \ne sign(\mu_j)
\end{equation}

\begin{figure}[ht]
	\centering
	\subfigure[Honest Gradient of CNN]{
		\includegraphics[width=0.45\columnwidth]{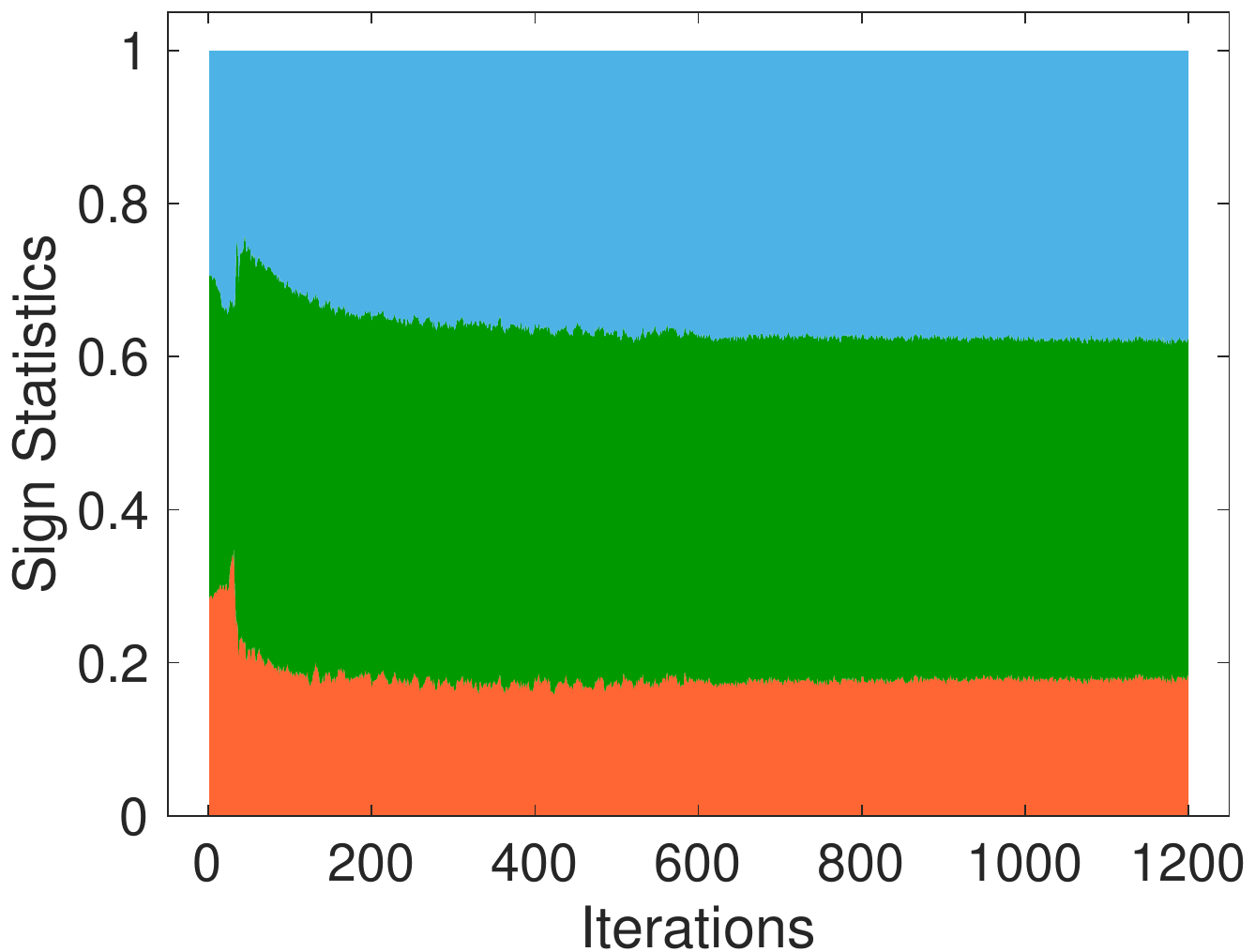}
	}
	%	\quad
	\subfigure[Malicious Gradient of CNN]{
		\includegraphics[width=0.45\columnwidth]{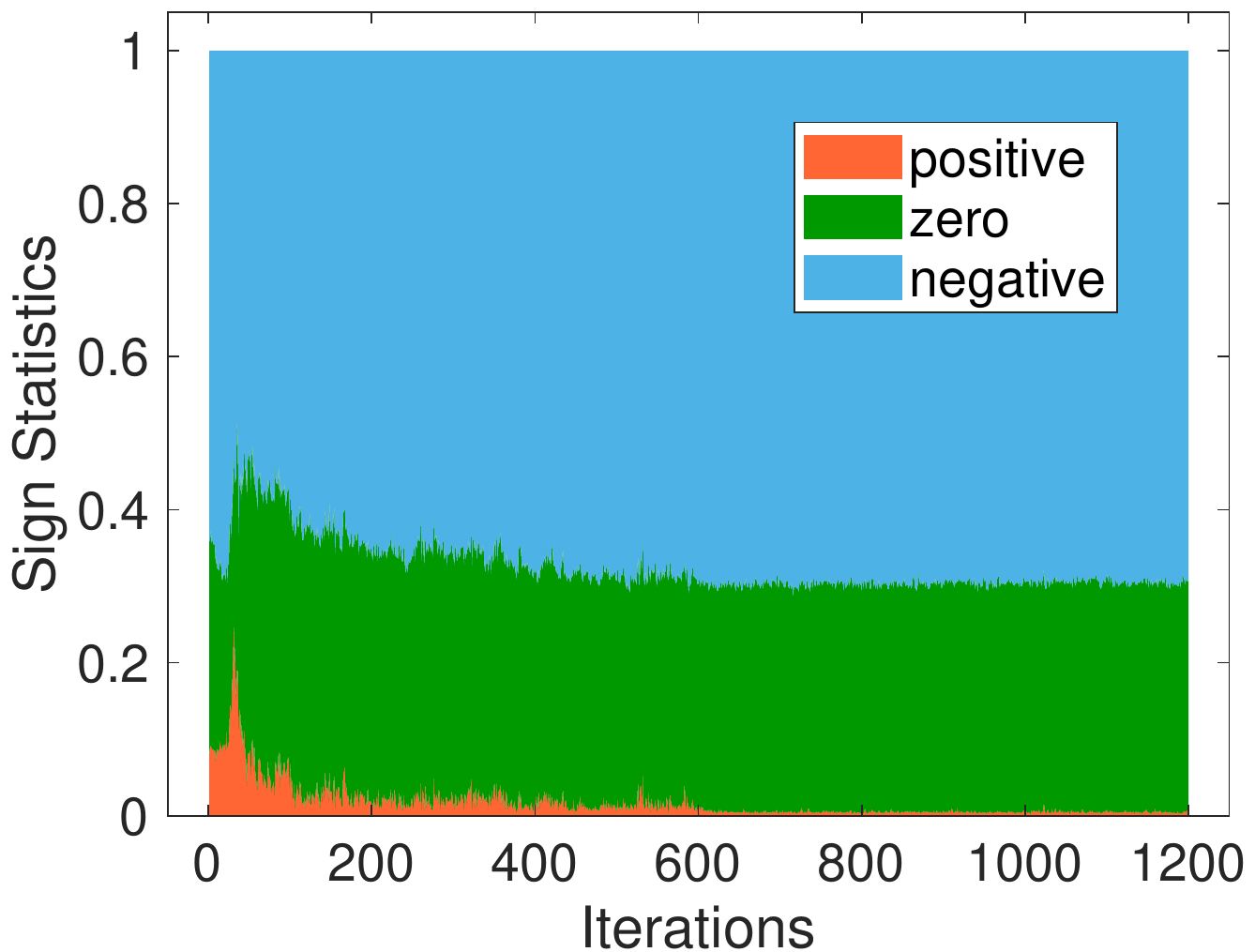}
	}
	%	\quad
	\subfigure[Honest Gradient of ResNet18]{
		\includegraphics[width=0.45\columnwidth]{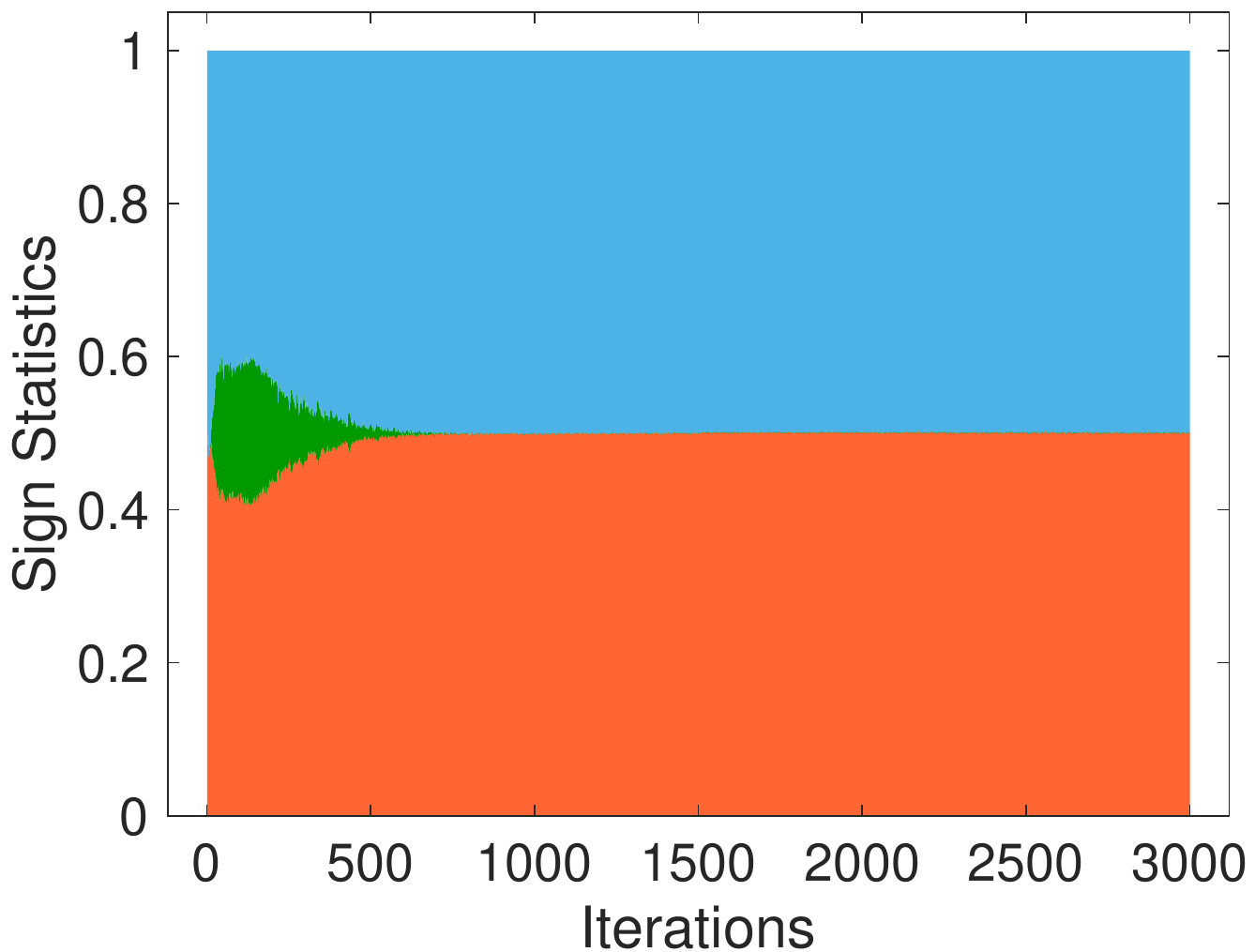}
	}
	%	\quad
	\subfigure[Malicious Gradient of ResNet18]{
		\includegraphics[width=0.45\columnwidth]{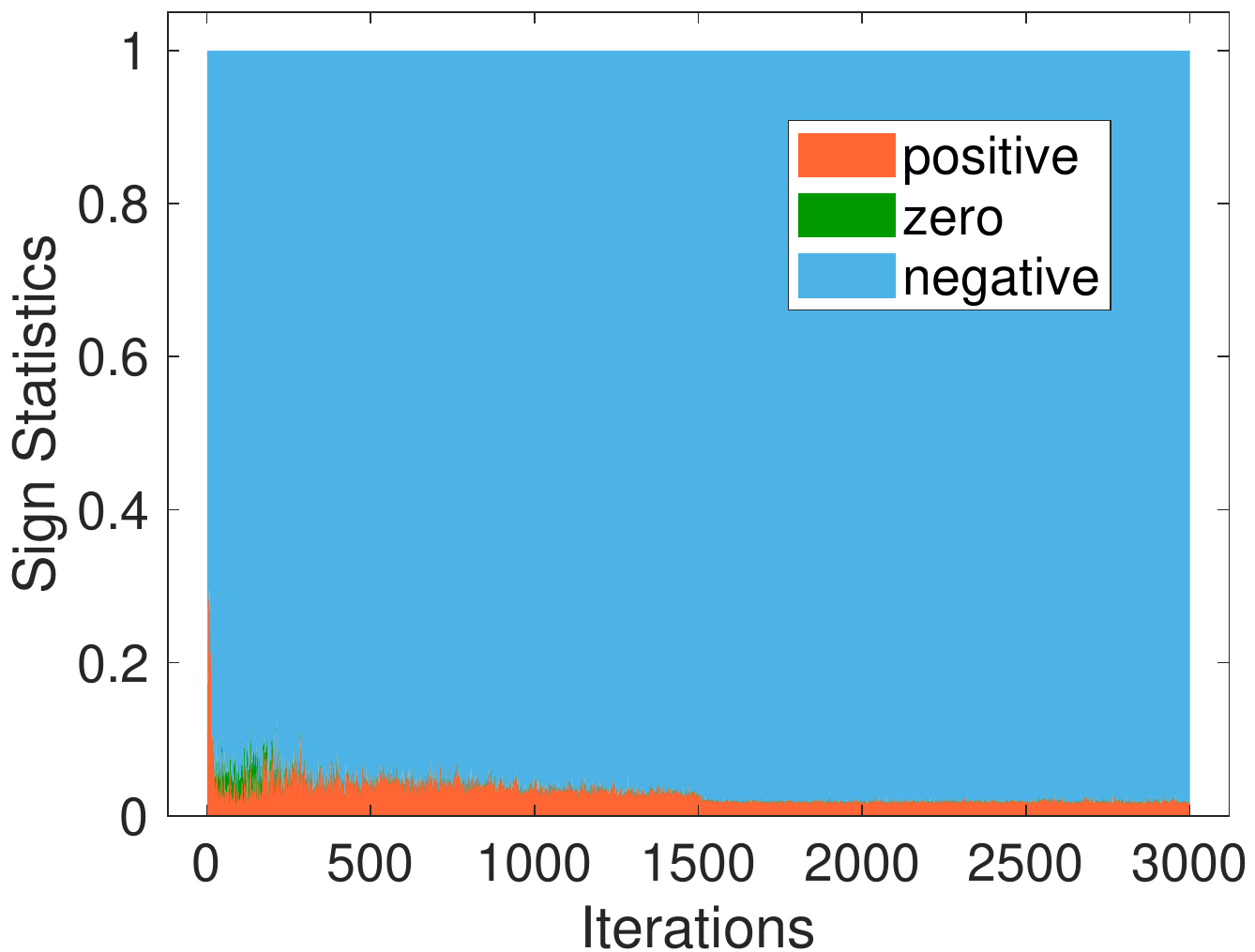}
	}
	\caption{Sign statistics of honest and malicious gradient. }
	\label{fig:sign_byz}
	\vspace{-3ex}
\end{figure}

Empirical results in \cite{BaruchBG19LIE} show that the coordinate-wise standard deviation turns out to be bigger than the corresponding mean, thus a small value of $z$ is enough to turn a large amount of positive elements into negative, leading to incorrect model updating. To verify this insight, we adopt the default training setting in Section~\ref{secExperimentSetup} to train a CNN on MNIST dataset and a ResNet-18 on CIFAR-10 dataset under no attacks. We calculate the averaged sign statistics across all clients as well as the sign statistics of a virtual gradient crafted by Eq. (\ref{eq:lie}) and plot them over iterations as in  Fig.~\ref{fig:sign_byz}, which convincingly supports our intuition. 

\vspace{1ex}
Next, we present the following Proposition~\ref{proposition:Safe-LIE} to explain why \textit{LIE} attack is hard to detect, where we compare the distance to averaged true gradient $\tilde{g}=\frac{1}{n}\sum_{i=1}^{n}g^{(i)}$ and similarity with $\tilde{g}$ for the malicious gradient and honest gradient, respectively.

\begin{proposition}
	For a distributed non-convex optimization problem $F(\mathbf{x})$ with $(n-m)$ benign workers and $m$ malicious workers conducting \textit{LIE} attack, suppose the data are IID and the gradient variance is bounded by $\sigma^2$. Given small enough $z$, then the distance between malicious gradient and true averaged gradient could be smaller than that of certain honest gradient:
	\begin{equation}
	\exists ~i, ~s.t. ~~ \mathbb{E}[\left\|g_m-\tilde{g}\right\|^2] < \mathbb{E}[\|g^{(i)}-\tilde{g}\|^2]
	\end{equation}
	and the cosine-similarity between malicious gradient and true averaged gradient could be bigger than that of certain honest gradient:
	\vspace{-1ex}
	\begin{equation}
	\exists ~i, ~s.t. ~~cos(g_m,\tilde{g}) > cos(g^{(i)},\tilde{g}) 
	\end{equation}
	\vspace{-2ex}
	\label{proposition:Safe-LIE}
\end{proposition}
\begin{proof}
	\vspace{-2ex}
	Detailed proof is in Appendix~\ref{appendix:prop1}.
\end{proof}

From the above analysis results, it can be concluded that the malicious gradient can be even ``safer'' when evaluated by Krum and Bulyan methods. Hence, it's difficult to detect the malicious gradient from the distance and cosine-similarity perspectives. Instead, checking the sign statistics is a novel and promising perspective to detect abnormal gradients. Our analysis is also valid for the recent proposed Min-Max/Min-Sum attacks in \cite{shejwalkar2021manipulating}.

\vspace{1ex}
\noindent \textbf{New Hybrid Attack.} In this work, we propose a type of hybrid attack called \textbf{ByzMean} attack, which makes the mean of gradients be arbitrary malicious gradient. More specifically, the malicious clients are divided into two sets, one set with $m_1$ clients chooses an arbitrary gradient vector $g_{m_1}=*$, and the other set with $m_2=m-m_1$ clients chooses the gradient vector $g_{m_2}$ such that the average of all gradients is exactly the $g_{m_1}$, which can be expressed as follows:
\begin{equation}
g_{m_1} = *, ~ g_{m_2}=\frac{(n-m_1)g_{m_1}-\sum_{i=m+1}^{n}g^{(i)}}{m_2}
\label{eq:byzMean}
\end{equation}
All existing attacks can be integrated into this ByzMean attack, making this hybrid attack even stronger than all single attacks. For example, we can set $g_{m_1}$ as random noise or the gradient crafted by {LIE} attack.

\section{Our SignGuard Framework}
\label{secFramework} 

In this section, we present formal problem formulation and introduce our SignGuard framework for Byzantine-robust federated learning. And some theoretical analysis on training convergence is also provided. 
%\vspace{-2ex}
\subsection{System Overview and Problem Setup}
Our federated learning system consists of a parameter server and a number of benign clients along with a small portion of Byzantine clients. We assume there exists an attacker or say adversary that aims at poisoning the global model and controls the Byzantine clients to perform malicious attacks. We first give out the following definitions of benign and Byzantine clients, along with the attacker's capability and defense goal.
\begin{definition}
	\textbf{(Benign Client)} A benign client always sends honest gradient to the server, which is an unbiased estimation of local true gradient at each iteration.
\end{definition}
\begin{definition}
	\textbf{(Byzantine Client)} A Byzantine client (also called corrupted client) may act maliciously and can send arbitrary message to the server.
\end{definition}

%\vspace{1ex}
\noindent \textbf{Threat Model.} Similar to the threat models in previous works \cite{BaruchBG19LIE,Fang20Local,shejwalkar2021manipulating}, we assume that there exists an attacker that controls some malicious clients to perform model poisoning attacks. Specially, we assume the attacker has full knowledge of all benign gradients and model parameters while the corrupted clients can also collude to conduct strong attacks. However, the attacker cannot corrupt the server and the proportion of malicious clients $\beta$ is less than half. 

\vspace{1ex}
\noindent \textbf{Defender's Capability:} As in previous studies \cite{Fang20Local,cao20FLTrust}, We consider the
defense is performed on the server-side. The parameter server does not have access to the raw training data on the clients, and the server does not know the exact number of malicious clients. However, the server has full access to the global model as well as the local model updates (i.e., local gradients) from all clients in each iteration. Specially, we further assume the received gradients are anonymous, which means the behavior of each client is untraceable. In consideration of privacy and security, we think this assumption is reasonable in some FL scenarios.

\vspace{1ex}
\noindent \textbf{Defense Goal:}
As mentioned in \cite{cao20FLTrust}, an ideal defense method should consider the following three aspects: Fidelity, Robustness, and Efficiency. We hope the defense method achieves Byzantine robustness against various malicious attacks without sacrificing the model accuracy. Moreover, the defense should be computationally cheap such that does not affect the overall training efficiency.

\vspace{1ex}
\noindent \textbf{Problem Formulation:}
We focus on federated learning on IID settings and then extend our algorithm into non-IID settings. We assume that training data are distributed over a number of clients in a network, and all clients jointly train a shared model based on disjoint local data. Mathematically, the underlying distributed optimization problem can be formalized as follows:
\begin{equation}\label{eq:objective}
\min_{\mathbf{x}\in R^d}{F(\mathbf{x})}=\frac{1}{n}\sum_{i=1}^{n}\mathbb{E}_{\xi_i\sim D_i}\left[F(\mathbf{x};\xi_i)\right]
\end{equation}
where $n$ is the total number of clients, $D_i$ denotes the local dataset of \textit{i}-th client and could have different distribution from other clients, and $F(\mathbf{x};\xi_i)$ denotes the local loss function given shared model parameters $\mathbf{x}$ and training data $\xi_i$ sampled from $D_i$. We make all clients initialize to the same point $\mathbf{x_0}$, then FedAvg \cite{mcmahan17} can be employed to solve the problem. At each iteration, the \textit{i}-th benign client draws $\xi_i$ from $D_i$, and computes local stochastic gradient with respect to global shared parameter $\mathbf{x}$, while Byzantine clients can send arbitrary gradient message:
\begin{equation}
\begin{aligned}
g_{t}^{(i)} = \begin{cases} 
\nabla F(\mathbf{x}_{t};\xi_i) ,&\text{if \textit{i}-th client is benign}
\\
   arbitrary , &\text{if \textit{i}-th client is Byzantine}
\end{cases}
\end{aligned}
\end{equation} 
The parameter server collects all the local gradients and employs robust gradient aggregation rule to get a global model update:
\begin{equation}
\mathbf{x_{t+1}} = \mathbf{x_{t}} -\eta_{t}\cdot \textsl{GAR}(\{g_{t}^{(i)}\}_{i=1}^{n})
\end{equation}
In synchronous settings with full client participation, the result will be broadcast to all clients to update their local models and start a new iteration. In a partial participation setting, the model update is finished in PS and the updated model will be sent to the selected clients for the next round. This process will repeat until the stop condition is satisfied.

To characterize the impact of Byzantine attacks, we define the following metric to  measure the effect of Byzantine attack by calculating the accuracy drop due to model poisoning:

\begin{definition}
	\textbf{(Attack Impact)} The impact of a specific attack is measured by the model accuracy drop compared to the baseline without the presence of any attack or defense.
\end{definition}
\vspace{-1ex}

\begin{figure*}[htbp]
	\centering
	{
		%\hspace{1ex}
		\includegraphics[width=2.0\columnwidth,clip=true]{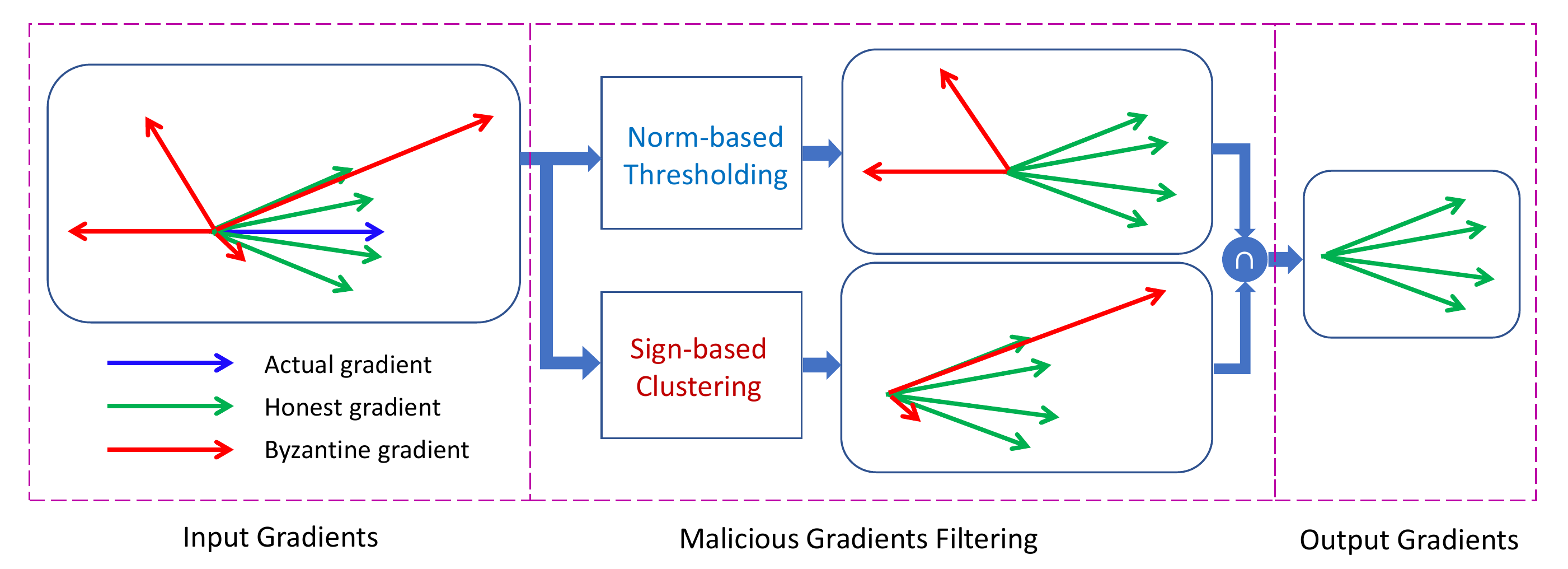}
	}	
	\vspace{-2ex}
	\caption{Illustration of the workflow of proposed SignGuard. The collected gradients are anonymous and sent into multiple filters in parallel, after which the intersection of multiple outputs are selected as trusted gradients. We use norm-based and sign-based filters in this paper.}
	%\vspace{-2ex}
	\label{fig:illustration}
\end{figure*}

\vspace{1ex}

\subsection{Our Proposed Solution}

The proposed SignGuard framework is described in Algorithm~\ref{alg1}-\ref{alg2} and the workflow is illustrated in Fig.~\ref{fig:illustration}. On a high level, we pay attention to the magnitude and direction of the received gradients. At each iteration, the collected gradients are sent into multiple filters, including norm-based thresholding filer and sign-based clustering filter. {Firstly}, for the norm-based filter, the median of gradient norms is utilized as reference norm as the median always lies in the benign set. Considering that small magnitudes of gradients do less harm to the training while a significantly large one is malicious, we will perform a loose lower threshold and a strict upper threshold. {Secondly}, for the sign-based clustering filter, we extract some statistics of gradients as features and use Mean-Shift \cite{meanshif} algorithm as an unsupervised clustering model with an adaptive number of cluster classes, while the cluster with the largest size is selected as the trusted set (if all malicious clients send the same attack vector, K-Means with two clusters will suffice). In this work, the proportions of positive, zero, and negative signs are computed as basic features, which are sufficient for a variety of attacks, including the LIE attack. 
\begin{algorithm}[t] 
	\setstretch{1}
	\caption{~SignGuard-based Robust Federated Learning} 
	\begin{algorithmic}[1] 
		\State \textbf{Input:} learning rate $\eta$, total iterations $T$, number of clients $n$
		\State \textbf{Initial:} $\mathbf{x}_0\in R^d$
		\For{$t=0, 1, ..., T-1$} 
		\State \textbf{On each client \textit{i} :}
		\State Sample a mini-batch of data to compute gradient  $\displaystyle g_{t}^{(i)}$
		%\vspace{1ex}
		\State Send $\displaystyle g_{t}^{(i)}$ to the parameter server
		\State Wait for global gradient $\tilde{g}_{t}$ from server
		\State Update local model:  $\displaystyle \mathbf{x}_{t+1}=\mathbf{x}_{t}-\eta\tilde{g}_{t}$ 
		\vspace{1ex}
		\State \textbf{On server:}
		\State Collect gradients from all clients
		%\vspace{1ex}
		\State Obtain global gradient: $\displaystyle\tilde{g}_{t}=SignGuard(\{g_{t}^{(i)}\}_{i=1}^{n})$
		\State Send $\tilde{g}_{t}$ to all clients
		\EndFor 
	\end{algorithmic} 
	\label{alg1}
\end{algorithm}

\begin{algorithm}[t] 
	\setstretch{1}
	\caption{~SignGuard Function} 
	\begin{algorithmic}[1] 
		\State \textbf{Input:} Set of received gradients $S_t=\{g_{t}^{(i)}\}_{i=1}^{n}$, lower and upper bound $L,R$ for gradient norm
	
		\State \textbf{Initial:} $S_1 = S_2 = \emptyset $
		\State \quad Get $l_2$-norm and element-wise sign of each gradient
		
		\State \textbf{Step 1:} Norm-based Filtering
		\State \quad Get the median of norm $M = med(\{\|g_{t}^{(i)}\|\}_{i=1}^{n})$
		\vspace{1ex}
		\State \quad Add the gradient that satisfies $L \leq  \dfrac{\|g_{t}^{(i)}\|}{M} \leq R $ into $S_1$
		
		\State \textbf{Step 2:} Sign-based Clustering
		\State \quad Randomly select a subset of gradient coordinates
		\State \quad Compute sign statistics on selected coordinates for each gradient as features
		\State \quad Train a Mean-Shift clustering model
		\State \quad Choose the cluster with most elements as $S_2$
		\State \textbf{Step 3:} Aggregation
		\State \quad Get trusted set: $S'_t=S_1 \cap S_2$ 
		\State \quad Get $\displaystyle\tilde{g}_{t}=\frac{1}{|S'_t|}\sum_{i\in S'_t}g_t^{(i)}\cdot \min\left(1,~{M}/{\|g_{t}^{(i)}\|}\right)$ 
		
		\State \textbf{Output:} Global gradient: $\displaystyle\tilde{g}_{t}$
	\end{algorithmic} 
	\label{alg2}
\end{algorithm}

However, those features only consider the overall statistics and lose sight of local properties. For example, when the amounts of positive and negative elements are approximate (e.g., ResNet-18), the naive sign statistics may be insufficient to detect the reversed gradients \cite{Rajput19Detox} or those well-crafted attacks that have similar sign statistics. To overcome this problem, we introduce randomized coordinate selection and add another similarity metric as an additional feature in our algorithm, such as cosine-similarity or Euclidean distance between each received gradient and a ``correct'' gradient. However, without the help of auxiliary data in PS, the ``correct'' gradient is not directly available. A practical way is to compute pairwise similarities between all the other gradients and take the median as the similarity with a ``correct'' gradient. Or more efficiently, just utilize the aggregated gradient from the previous iteration as the ``correct'' gradient. Intuitively, it is promising to distinguish those irrelevant gradients and helps to improve the robustness of anomaly detection. However, as shown in Section~\ref{secAttackAnalysis}, the Euclidean distance or cosine-similarity metrics are not reliable for the state-of-the-art attacks, and even affect the judgment of SignGuard as we found in experiments. In this work, the plain ``SignGuard" only uses sign statistics in default, and the enhanced variants that add the cosine-similarity feature or Euclidean distance feature are called ``SignGuard-Sim" and ``SignGuard-Dist", respectively. We will provide some comparative results of them. How to design a more reliable similarity metric is left as an open problem for future work.

After the filtering process, the server eventually selects the intersection of multiple filter outputs as the trusted set and obtains a global gradient by robust aggregation, e.g. trimmed-mean. In this work, we use the mean aggregation with norm clipping, where the clipping bound is selected as the median value of gradient norms. It is worth noting that a small fraction of honest gradients could also be filtered out, especially in the non-IID settings, depending on the variance of honest gradients and the distance to malicious gradients.

\subsection{Convergence Analysis}

To conduct convergence analysis, we also make the following basic assumption, which is commonly used in the literature \cite{yu2019on,bottou2018optim,Karimireddy19error_fix} for convergence analysis of distributed optimization. 
%Missing proofs can be referred to the full version in \cite{xu2021signguard}.
\begin{assumption}
	Assume that problem (\ref{eq:objective}) satisfies:
	
%	\vspace{2ex}
	\textbf{1. Smoothness}: The objective function $F(\cdot)$ is smooth with Lipschitz constant $L>0$, which means $\forall \mathbf{x}, \forall \mathbf{y},~\left\| \nabla F(\mathbf{x})-\nabla F(\mathbf{y})\right\| \leq L\left\| \mathbf{x}-\mathbf{y}\right\|$.
	It implies that:
	\begin{equation}
	F(\mathbf{x})-F(\mathbf{y}) \leq \nabla F(\mathbf{x})^{T}(\mathbf{y}-\mathbf{x})+\frac{L}{2}\left\| \mathbf{x}-\mathbf{y}\right\|^2  \notag
	\end{equation}
	
	\textbf{2. Unbiased local gradient}: For each worker with local data, the stochastic gradient is locally unbiased:
	\begin{equation}
	\mathbb{E}_{\xi_i\sim D_i}\left[\nabla F(\mathbf{x};\xi_i)\right] = \nabla F_i(\mathbf{x}) \notag
	\end{equation}
	
	\textbf{3. Bounded variances}: The stochastic gradient of each worker has a bounded variance uniformly, satisfying:
	\begin{equation}
	\mathbb{E}_{\xi_i\sim D_i}[\left\|\nabla F(\mathbf{x};\xi_i)-\nabla F_i(\mathbf{x})\right\|^2] \leq \sigma^2 \notag
	\end{equation}
	and the deviation between local and global gradient satisfies:
	\begin{equation}
	\left\|\nabla F_i(\mathbf{x})-\nabla F(\mathbf{x})\right\|^2 \leq \kappa^2 \notag
	\end{equation}
	
	\label{as:1}
\end{assumption}
%\vspace{-2ex}

For the SignGuard framework, the trusted gradients attained by filters may still contain a part of malicious gradients. In this case, any gradient aggregation rule necessarily results in an error to the averaged honest gradient \cite{LaiRV16agnostic,Karimireddy20history}. Here we make another assumption on the capability of the aggregation:
\begin{assumption}
	For problem (\ref{eq:objective}) with $(1 - \beta)n$ benign clients (denoted by $\mathcal{G}$) and $\beta n$ Byzantine clients, suppose that at most $\delta n$ Byzantine clients can circumvent SignGuard at each iteration. We assume that there exist positive constants $c$ and $b$ such that the output $\hat{g}_t$ of SignGuard satisfies:
	\begin{equation}
	\begin{aligned}
	&\textbf{1. Bounded Bias:}~~\left[\mathbb{E}\left\|\hat{g}_t-\bar{g}_t\right\|\right]^2
	\leq c{\delta}\sup_{i,j\in \mathcal{G}}\mathbb{E}[\|{g_t^{(i)}}-{g_t^{(j)}}\|^2]\\
	&\textbf{2. Bounded Variance:}~~ \text{var}\left\|\hat{g}_t\right\|
	\leq b^2 \notag
	\end{aligned}	
	\end{equation}
	where $\bar{g}_t=\frac{1}{|\mathcal{G}|}\sum_{i\in \mathcal{G}}g_t^{(i)}$ and $0\le \delta < \beta<0.5~$.
	\label{as:2}
\end{assumption}

\begin{remark}
When $\delta=0$, it's possible to exactly recover the averaged honest gradient. For most aggregation rules such as Krum, the output is deterministic and thus has $b^2=0$. For clustering-based rules, the output is randomized and could have negligible variance if the clustering algorithm is robust. The bounded bias assumption is reasonable since we perform norm clipping before aggregation.
\end{remark}
\vspace{-1ex}
When $\beta n$ Byzantine clients exist and act maliciously, the desired gradient aggregation result is the average of $(1 - \beta)n$ honest gradients, which still has a deviation to the global gradient of no attack setting. We give the following lemma to characterize the deviation:
\vspace{-1ex}
\begin{lemma}
	Suppose the training data are non-IID under Assumption 1, then the deviation between averaged gradient of $(1-\beta)n$ clients $\bar{g}$ and the true global gradient $\nabla F(\mathbf{x})$ can be characterized as follows:
	\begin{equation}
	\mathbb{E}\left[\left\|\bar{g}-\nabla F(\mathbf{x})\right\|^2\right]
	\leq \frac{\beta^2\kappa^2}{(1-\beta)^2}+\frac{\sigma^2}{(1-\beta)n} 
	\end{equation}
\end{lemma}
\begin{proof}
	Detailed proof is in Appendix \ref{appendix:lemm1}.
\end{proof}

%\vspace{-1ex}
Given the above assumptions and lemma, extending the analysis techniques in \cite{bottou2018optim,yu2019on, Karimireddy19error_fix, Karimireddy20history}, now we can characterize the convergence of SignGuard by the following theorem. 
\vspace{-1ex}
\begin{theorem}
	For problem (\ref{eq:objective}) under Assumption 1, suppose the SignGuard satisfying Assumption 2 is employed with a fixed learning rate $\eta \le (2-\sqrt{\delta}-2\beta)/(4L)$ and $F^*=\min_{\mathbf{x}}F(\mathbf{x})$, then we have the following result:
	\begin{equation}
	\dfrac{1}{T}\sum_{t=0}^{T-1}\mathbb{E}[\left\|\nabla F(\mathbf{x}_t)\right\|^2] \leq \frac{2(F(\mathbf{x}_0)-F^*)}{\eta T}+2L\eta\Delta_1 + \Delta_2 \notag
	\end{equation}
	where the constant terms are $\Delta_1=4c\delta(\sigma^2+\kappa^2)+2b^2+\frac{2\beta^2\kappa^2}{(1-\beta)^2}+\frac{2\sigma^2}{(1-\beta)n}$ and $\Delta_2=4c\sqrt{\delta}(\sigma^2+\kappa^2)+\frac{\beta\kappa^2}{(1-\beta)^2}$.
	\label{theorem:signGuard}
\end{theorem}
\begin{proof}
	Detailed proof is in Appendix \ref{append:theorem1}.
\end{proof}

\begin{remark}
	The terms $\Delta_1$ and $\Delta_2$ arise from the existence of Byzantine clients and are influenced by the capability of aggregation rule. When no Byzantine client exists ($\beta=0$ and thus $\delta=0$), we have $\Delta_2=0$ and the convergence is guaranteed with a sufficiently small learning rate. If Byzantine clients exist ($\beta>0$), even the defender is capable to remove all malicious gradients ($\delta=0$), we still have $\Delta_2>0$ due to non-IID data and may result in some model accuracy gaps to benchmark results. 

\end{remark}

\section{Experimental Setup}\label{secExperimentSetup}
The proposed SignGuard framework is evaluated on various datasets for image and text classification tasks. We mainly implement the learning tasks in the IID fashion, and investigate the performance of different defenses in the non-IID settings as well. The models that trained under no attack and no defense are used as benchmarks.

\subsection{Datasets and Models}

\noindent \textbf{MNIST.} MNIST is a 10-class digit image classification dataset, which consists of 60,000 training samples and 10,000 test samples, and each sample is a grayscale image of size 28 × 28. For MNIST, we construct a convolutional neural network (CNN) with 3 convolutional layers and 2 fully-connected layers as the global model.

\noindent \textbf{Fashion-MNIST.} Fashion-MNIST \cite{xiao17fmnist} is a clothing image classification dataset, which has the same image size and structure of training and testing splits as MNIST, and we use the same CNN model as in MNIST.

\noindent \textbf{CIFAR-10.} CIFAR-10 \cite{cifar10/100} is a well-known color image classification dataset with 60,000 32 × 32 RGB images in 10 classes, including 50,000 training samples and 10,000 test samples. We use ResNet-18 \cite{he2016residual} as the global models\footnote{We use open-source implementation of ResNet-18, which is available at https://github.com/kuangliu/pytorch-cifar}.

\noindent \textbf{AG-News.} AG-News is a 4-class topic classification dataset. Each class contains 30,000 training samples and 1,900 testing samples. The total number of training samples is 120,000 and 7,600 for test. We use the TextRNN with a two-layer bi-directional LSTM network \cite{LiuQH16textRNN} as the global model.

\subsection{Evaluated Attacks}
We consider various popular model poisoning attacks:

\textbf{Random Attack.} The Byzantine clients send gradients with randomized values that generated by a multi-dimensional Gaussian distribution ${N}(\mu,\sigma^2 \textbf{I})$. In our experiments, we take $\mu = (0,...,0) \in \mathbb{R}^d\ $and $\sigma=0.5$ to conduct random attacks.

\textbf{Noise Attack.} The Byzantine clients send noise perturbed gradients that generated by adding Gaussian noise into honest gradients: $ g_{m} = g_{b} + {N}(\mu,\sigma^2 \textbf{I}) $. We take the same Gaussian distribution parameters as random attack.

\textbf{Sign-Flipping.} The Byzantine clients send reversed gradients without scaling: $ g_{m} = -g_{b}$. This is a special case of reverse gradient attack \cite{Rajput19Detox,Xie19Empires}.

\textbf{Label-Flipping.} The Byzantine clients flip the local sample labels during the training process to generate faulty gradients. This is also a type of data poisoning attack. In particular, the label of each training sample in Byzantine clients is flipped from $l$ to $C-1-l$, where $C$ is the total categories of labels and $l\in \{0,1,\cdots,C-1\}$.

\textbf{Little is Enough.} As in \cite{BaruchBG19LIE}, the Byzantine clients send malicious gradient vector with elements crafted as Eq.~(\ref{eq:lie}). We set $z=0.3$ for default training settings in our experiments.

\textbf{ByzMean Attack.} As proposed in Section~\ref{secAttackAnalysis}, we set $m_1=\lfloor 0.5m \rfloor$ and $m_2=m-m_1$, and set $g_{m_1}$ as LIE attack.

\textbf{Min-Max/Min-Sum.} As in \cite{shejwalkar2021manipulating}, the malicious gradient is a perturbed version of the benign aggregate as Eq.~(\ref{eq:gstd}), where $\nabla^p$ is a perturbation vector and $\gamma$ is a scaling coefficient, and those two attacks are formulated in Eq.~(\ref{eq:minmax})-(\ref{eq:minsum}). The first Min-Max attack ensures that the malicious gradients lie close to the clique of the benign gradients, while the Min-Sum attack ensures that the sum of squared distances of the malicious gradient from all the benign gradients is upper bounded by the sum of squared distances of any benign gradient from the other benign gradients. To maximize the attack impact, all malicious gradients keep the same. By default, we choose $\nabla^p$ as $-std(g^{\{i\in [n]\}})$, i.e., the inverse standard deviation.
\begin{equation}
g_m = f_{avg}(g^{\{i\in [n]\}})+\gamma \nabla^p
\label{eq:gstd}
\end{equation}
\begin{equation}
	\mathop{\arg\max}\limits_{\gamma} ~ \mathop{\max}\limits_{i\in [n]}\|g_m-g^{(i)}\|\leq \mathop{\max}\limits_{i,j\in [n]}\|g^{(i)}-g^{(j)}\|
	\label{eq:minmax}
\end{equation}
\begin{equation}
	\mathop{\arg\max}\limits_{\gamma} ~ \mathop{\sum}\limits_{i\in [n]}\|g_m-g^{(i)}\|^2\leq \mathop{\max}\limits_{i\in [n]}\mathop{\sum}\limits_{j\in [n]}\|g^{(i)}-g^{(j)}\|^2
	\label{eq:minsum}
\end{equation}

\subsection{Training Settings}
By default, we consider a FL setup with $n = 50$ clients, 20\% of which are Byzantine nodes, and the training data are IID among clients. To verify the resilience and robustness, we will also evaluate the impact of different fractions of malicious clients for different attacks and defenses. Furthermore, our approach will also be evaluated in realistic non-IID settings. In all experiments, we set the lower and upper bounds of gradient norm as $L = 0.1$ and $R = 3.0$, and randomly select 10\% of coordinates to compute sign statistics in our SignGuard-based algorithms. Each training algorithm is run for 60 epochs for MNIST/Fashion-MNIST/AG-News and 160 epochs for CIFAR-10. The number of local iteration is set to 1 and momentum is employed with the parameter of 0.9, and the weight decay is set to 0.0005. 

\section{Evaluation Results}\label{secExperimentResult}

\begin{table*}[ht] \scriptsize
	\centering
	\caption{Comparison of defenses under various model poisoning attacks}  
	\label{tab:main_result_iid} 
	
	\newcommand{\tabincell}[2]{\begin{tabular}{@{}#1@{}}#2\end{tabular}} 
	\begin{tabular}{| c | c | p{1.08cm}<{\centering} | p{1.08cm}<{\centering} | p{1.08cm}<{\centering} | c | p{1.08cm}<{\centering} | p{1.08cm}<{\centering} | p{1.08cm}<{\centering} | p{1.08cm}<{\centering} | p{1.08cm}<{\centering} |}
		\hline
		\multirow{2}{*}{\tabincell{c} {Dataset\\(Model)}} & 
		\multirow{2}{*}{GAR} & 
		\multirow{2}{*}{No Attack} & 
		\multicolumn{3}{c|}{Simple Attacks}&
		\multicolumn{5}{c|}{State-of-the-art Attacks}\\
		\cline{4-5} 
		\cline{5-6} 
		\cline{6-7}
		\cline{7-8}
		\cline{8-9} 
		\cline{9-10}  
		\cline{10-11}    
		& & & {Random} & {Noise} & {Label-flip} & {ByzMean} & {Sign-flip} & {LIE} & {Min-Max} & {Min-Sum}\\
		\hline
		
		\multirow{10}{*}{\tabincell{c} {MNIST\\(CNN)}}&
		Mean & 99.23 & 84.84 & 90.48 & 99.05 & 31.98 & 98.42 & 84.49 & 68.89 & 34.46 \\
		& TrMean & 98.23 & 98.63 & 98.53 & 95.31 & 58.87 & 98.44 & 94.50 & 34.48 & 43.89\\
		& Median & 97.46 & 94.18 & 97.45 & 93.84 & 40.04 & 97.73 & 74.37 & 26.11 & 38.13 \\
		& GeoMed & 93.21 & 82.77 & 78.68 & 86.20 & 45.02 & 74.78 & 34.37 & 15.62 & 20.53 \\
		& Multi-Krum & 99.20 & 98.98 & 99.11 & 99.06 & 83.26 & 98.82 & 90.04 & 52.77 & 27.27 \\
		& Bulyan & 99.10 & 99.17 & {99.12} & 99.15 & 98.58 & 98.81 & 98.86 & 52.45 & 51.95 \\
		& DnC & 99.09 & 99.07 & 99.08 & {99.17} & 82.25 & 98.73 & {99.12} & 98.97 & 81.04 \\
		& SignGuard & 99.11 & {99.09} & {98.97} & \textbf{99.18} & \textbf{99.02} & \textbf{99.13} & 99.15 & \textbf{99.18} & {99.15} \\
		& SignGuard-Sim & 99.16 & \textbf{99.18} & 99.16 & 99.07 & {98.91} & {99.06} & \textbf{99.22} & {99.08} & {99.13} \\
		& SignGuard-Dist & 98.95 & 99.05 & \textbf{99.18} & 99.11 & 98.93 & 98.86 & 98.96 & 99.01 & \textbf{99.19} \\
		\hline \hline
		
		\multirow{10}{*}{\tabincell{c} {Fashion-MNIST\\(CNN)}}&
		Mean & 89.51 & 69.88 & 31.83 & 89.37 & 16.31 & 86.68 & 79.78 & 47.73 & 45.12 \\
		& TrMean & 87.02 & 87.81 & 87.45 & 79.58 & 62.66 & 87.45 & 54.28 & 45.71 & 42.96 \\
		& Median & 80.77 & 82.96 & 82.59 & 77.41 & 47.46 & 82.52 & 45.14 & 47.43 & 50.83 \\
		& GeoMed & 76.51 & 79.96 & 78.93 & 78.16 & 40.51 & 70.65 & 10.00 & 73.75 & 66.63 \\
		& Multi-Krum & 87.89 & 89.12 & 88.94 & 89.27 & 69.95 & 87.59 & 72.22 & 40.08 & 47.36 \\
		& Bulyan & 88.80 & 89.31 & 89.32 & 89.21 & 88.72 & 87.52 & 88.64 & 59.65 & 43.63 \\
		& DnC & 89.21 & 88.89 & 88.14 & 88.85 & 70.15 & 87.58 & 71.82 & 88.43 & 88.94 \\
		& SignGuard & 89.48 & \textbf{89.34} & \textbf{89.32} & 89.12 & 89.35 & 88.69 & 89.34 & \textbf{89.48} & \textbf{88.51} \\
		& SignGuard-Sim & 89.43 & {89.24} & 89.21 & \textbf{89.33} & {89.28} & {89.08} & \textbf{89.36} & {89.04} & {88.18}\\
		& SignGuard-Dist & 89.37 & 88.87 & 89.30 & 89.31 & \textbf{89.39} & \textbf{89.21} & \textbf{89.36} & 89.34 & 88.38 \\
		\hline \hline
		
		\multirow{10}{*}{\tabincell{c} {CIFAR-10\\(ResNet-18)}}&
		Mean & 93.16 & 44.53 & 46.34 & {91.98} & 17.18 & 79.63 & 55.86 & 23.84 & 18.17 \\
		& TrMean & 93.15 & 89.61 & 89.47 & 85.15 & {30.13} & 85.54 & 43.76 & 24.81 & 23.36 \\
		& Median & 74.18 & 68.27 & 71.42 & 71.19 & 23.47 & 70.75 & 27.35 & 20.46 & 22.74 \\
		& GeoMed & 65.62 & 70.41 & 69.35 & 70.76 & 24.86 & 67.82 & 23.55 & 50.36 & 45.23 \\
		& Multi-Krum & 93.14 & \textbf{92.88} & \textbf{92.81} & 92.26 & 50.41 & 92.36 & 42.58 & 21.17 & 38.24 \\
		& Bulyan & 92.78 & 91.87 & 92.47 & 92.24 & 81.33 & 90.12 & 74.52 & 29.87 & 37.79 \\
		& DnC & 92.73 & 88.01 & 88.25 & 92.05 & 36.56 & 84.76 & 47.37 & 52.94 & 35.36 \\
		& SignGuard & 93.03 & \textbf{92.78} &\textbf{92.52}  & 92.28 & \textbf{92.46} & 88.61 & \textbf{92.93} & 92.56 & {92.47} \\
		& SignGuard-Sim & 93.19 & 92.51 & 91.38  & 92.26 & {92.26} & \textbf{92.48} & 92.62 & {92.63} & 92.75 \\
		& SignGuard-Dist & 92.76 & 92.64 & 92.26 & \textbf{92.51} & 92.42 & 91.69 & 92.36 & \textbf{92.82} & \textbf{92.93} \\
		\hline \hline

		\multirow{10}{*}{\tabincell{c} {AG-News\\(TextRNN)}}&
		Mean & 89.36 & 28.18 & 28.41 & 86.72 & 25.05 & 84.18 & 79.34 & 27.32 & 25.24 \\
		& TrMean & 87.57 & 88.33 & 88.72 & 85.50 & 37.51 & 84.84 & 66.95 & 30.05 & 30.28 \\
		& Median & 84.57 & 84.52 & 84.59 & 82.08 & 28.99 & 81.10 & 32.39 & 30.28 & 29.71 \\
		& GeoMed & 82.38 & 77.63 & 77.18 & 78.42 & 27.36 & 81.64 & 31.57 & 74.82 & 71.48 \\
		& Multi-Krum & 88.86 & 89.18 & 89.22 & 86.89 & 68.53 & \textbf{87.42} & 72.98 & 53.51 & 32.46 \\
		& Bulyan & 88.22 & 88.86 & 88.93 & 85.54 & 85.80 & 86.55 & 85.49 & 47.76 & 51.25 \\
		& DnC & 89.13 & 86.42 & 86.28 & 86.72 & 31.47 & 86.30 & 76.58 & 88.45 & 89.05 \\
%		\cline{2-12}
		& SignGuard  & 89.29 & \textbf{89.22} & 89.23 & 86.78 & \textbf{89.24} & 86.53 & 89.26 & 89.23 & 89.27 \\
		& SignGuard-Sim & 89.24 & 89.13 & \textbf{89.29} & 87.05 & 89.36 & 86.76 & \textbf{89.33} & \textbf{89.27} & \textbf{89.37}\\
		& SignGuard-Dist & 89.23 & 89.16 & 89.23 & \textbf{87.25} & 89.31 & \textbf{87.30} & 89.17 & 89.22 & 89.35 \\
		\hline
	\end{tabular}
\end{table*}

\begin{figure*}[h]
	\centering
	\subfigure[CNN trained on Fashion-MNIST]{
		\includegraphics[width=2.0\columnwidth]{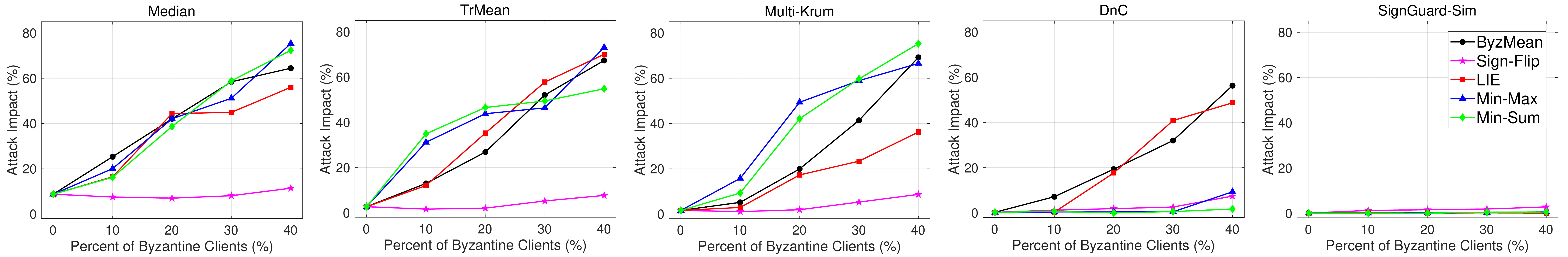}
	}
	%	\quad
	\subfigure[ResNet-18 trained on CIFAR-10]{
		\includegraphics[width=2.0\columnwidth]{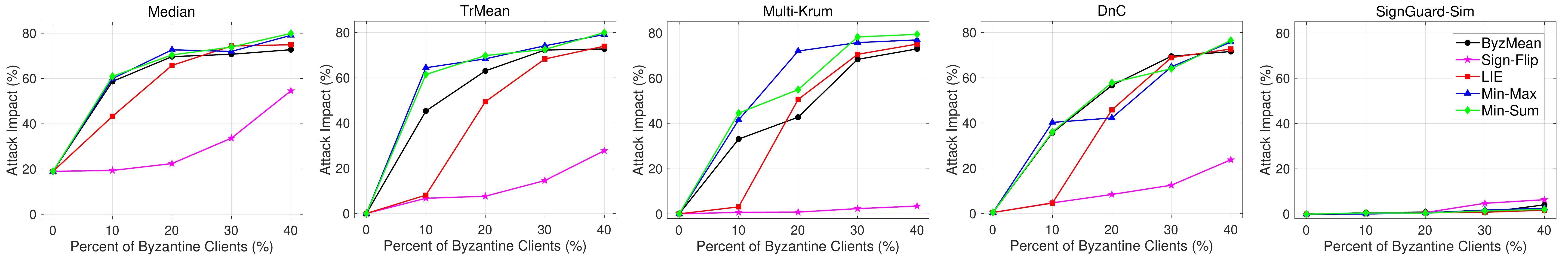}
	}
	%	\quad
	\caption{Accuracy drop comparison under various attacks and different percentage of Byzantine clients. SignGuard-Sim has the smallest gap to the baseline. }
	\label{fig:acc_byznum}
\end{figure*}

In this section, we conduct extensive experiments with various attack-defense pairs on both IID and non-IID data settings. We compare our methods with several existing defense methods, including TrMean, Median, GeoMed, Multi-Krum, Bulyan and DnC. The numerical results demonstrate the efficacy and superiority of our proposed SignGuard framework.

\subsection{Main Results in IID Settings}
The main results of best-achieved test accuracy during the training process under different attack and defense methods in the IID settings are collected in Table~\ref{tab:main_result_iid}. The results of naive \textit{Mean} aggregation under \textit{No Attack} are used as benchmarks. Notice that we favor other defenses by assuming the defense algorithms know the fraction of Byzantine clients, which is somewhat unrealistic but intrinsically required by existing defenses. However, we do not rely on the Byzantine fraction information in our SignGuard framework, which is an important advantage over existing methods.

\vspace{1ex}
\noindent \textbf{Performance comparison.} Test results on four datasets consistently demonstrate that our SignGuard-type methods can leverage the power of sign statistics and similarity features to filter out most malicious gradients and achieve competitive test accuracy as general SGD under no attack. Consistent with original papers \cite{BaruchBG19LIE,shejwalkar2021manipulating}, the state-of-the-art attacks, such as LIE and Min-Max/Min-Sum, can circumvent the median-based and distance-based defenses, preventing successful model training. Take the results of Multi-Krum on ResNet-18 as an example, it can be seen that when no attack is performed, Multi-Krum has a negligible accuracy drop (less than 0.1\%). However, the best test accuracy drops to 42.58\% under LIE attack and even less than 40\% under Min-Max/Min-Sum attacks. Similar phenomena can also be found in model training under TrMean, Median, and Bulyan methods. Besides, even under no attack, the Median and GeoMed methods are only effective in simple tasks, such as CNN for digit classification on MNIST and TextRNN for text classification on AG-News. When applied to complicated model training, such as ResNet-18 on CIFAR-10, those two methods have high convergence error and result in significant model degradation. While Muti-Krum and Bulyan suffer from well-crafted attacks, they perform well on naive attacks and even better than our plain SignGuard in mitigating random noise and sign-flipping attack. Though the DnC method has extraordinary effectiveness under many attacks, we found it is unstable during training and can be easily broken by our proposed ByzMean attack. In contrast, our proposed SignGuard-type methods are able to distinguish most of those well-crafted malicious gradients and achieve satisfactory model accuracy under various types of attacks. It is worth noting that our plain SignGuard already attains high robustness and fidelity, and the cosine-similarity/distance can further improve the defense performance in some cases, e.g., mitigating the sign-flipping attack. Besides, considering that the local data of Byzantine clients also contribute to the global model when no attack is performed, it's not surprising to see that even the best defense against Byzantine attack will still result in a small gap to the benchmark results.

We also report the average selected rate of both benign and Byzantine gradients during the training process of ResNet-18 in Table~\ref{tab:rate_iid}. We notice that the SignGuard-type methods inevitably exclude part of honest gradients, and select some malicious gradients under the sign-flipping attack. The reason lies in the fact that the proportions of positive and negative elements in normal gradient are approximate for ResNet-18. We also notice that although SignGuard-Sim is the most critical one and only selects less than 80\% honest gradients during training, it is resilient to various kinds of attacks and still achieves high accuracy results.

\begin{table}[ht] \scriptsize
	\centering
	\caption{Selected Rate of Honest and Malicious Gradients}  
	\label{tab:rate_iid} 
	
	\newcommand{\tabincell}[2]{\begin{tabular}{@{}#1@{}}#2\end{tabular}} 
	\begin{tabular}{| c | c | c | c | c | c | c |}
		\hline
		\multirow{2}{*}{\tabincell{c} {\textbf{Attack}}} & 
		\multicolumn{2}{c|}{\textbf{SignGuard}} &
		\multicolumn{2}{c|}{\textbf{SignGuard-Sim}} &
		\multicolumn{2}{c|}{\textbf{SignGuard-Dist}} \\
		\cline{2-3} 
		\cline{3-4} 
		\cline{4-5}
		\cline{5-6}
		\cline{6-7}
		& {H} & {M} & {H} & {M} & {H} & {M}\\
		\hline
		ByzMean &  0.9625 & 0 & 0.7791 & 0 & 0.9272 & 0.0003\\
		\hline
		Sign-flip & 0.6870 & 0.3908 & 0.7639 & 0.0981 & 0.7570 & 0.2440\\
		\hline
		LIE &  0.9532 & 0 & 0.7727 & 0 & 0.9151 & 0\\
		\hline
		Min-Max &  0.9650 & 0 & 0.7866 & 0.0003 & 0.9105 & 0.0009\\
		\hline
		Min-Sum &  0.9640 & 0 & 0.7752 & 0 & 0.9111 & 0\\
		\hline
	\end{tabular}
\end{table}

\vspace{1ex}
\noindent \textbf{Percentage of Byzantine clients.} We also evaluate the performance of signGuard-Sim with different percentages of Byzantine clients. In this part, we conduct experiments of CNN trained on the Fashion-MNIST dataset and ResNet-18 trained on CIFAR-10 dataset. We keep the total number of clients be 50 and vary the fraction of Byzantine clients from 10\% to 40\% to study the impact of Byzantine percentage for different defenses. We use the default training settings, and experiments are conducted under various state-of-the-art attacks. Particularly, we compare the results of SignGuard-Sim with Median, TrMean, Multi-Krum, and DnC as shown in Fig.~\ref{fig:acc_byznum}. It can be seen that our approach can effectively filter out malicious gradients and result in a slight accuracy drop regardless of the high percentage of Byzantine clients, while other defense algorithms suffer much more attack impact with the increasing percentage of Byzantine clients. In particular, we also find that Multi-Krum can mitigate sign-flipping attack well in ResNet-18 training, possibly because the exact percentage of Byzantine clients is provided to the Multi-Krum algorithm.

\begin{figure}[ht]
	\centering
	\subfigure[CNN on Fashion-MNIST]{
		\includegraphics[width=0.46\columnwidth]{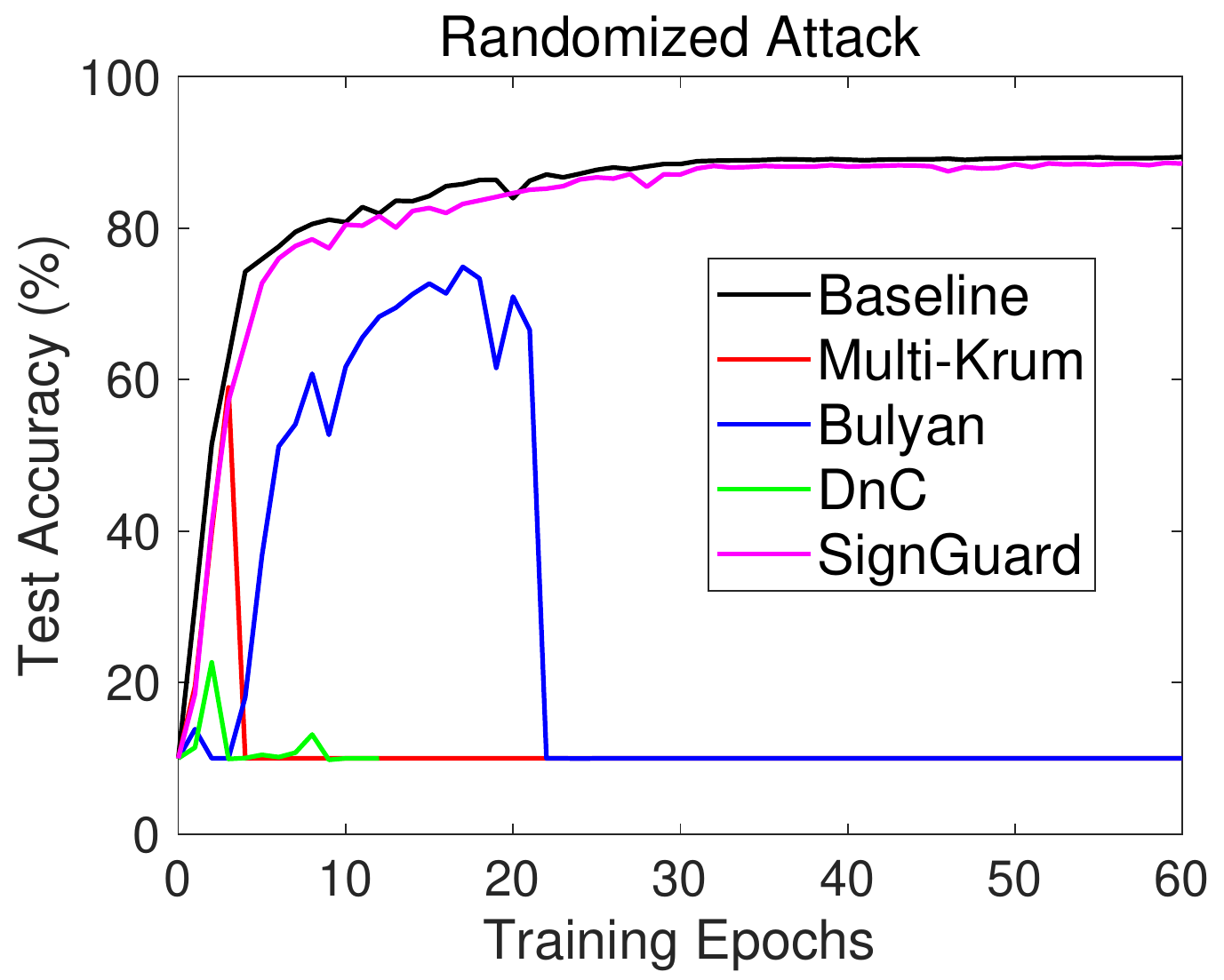}
	}
	%	\quad
	\subfigure[ResNet-18 on CIFAR-10]{
		\includegraphics[width=0.46\columnwidth]{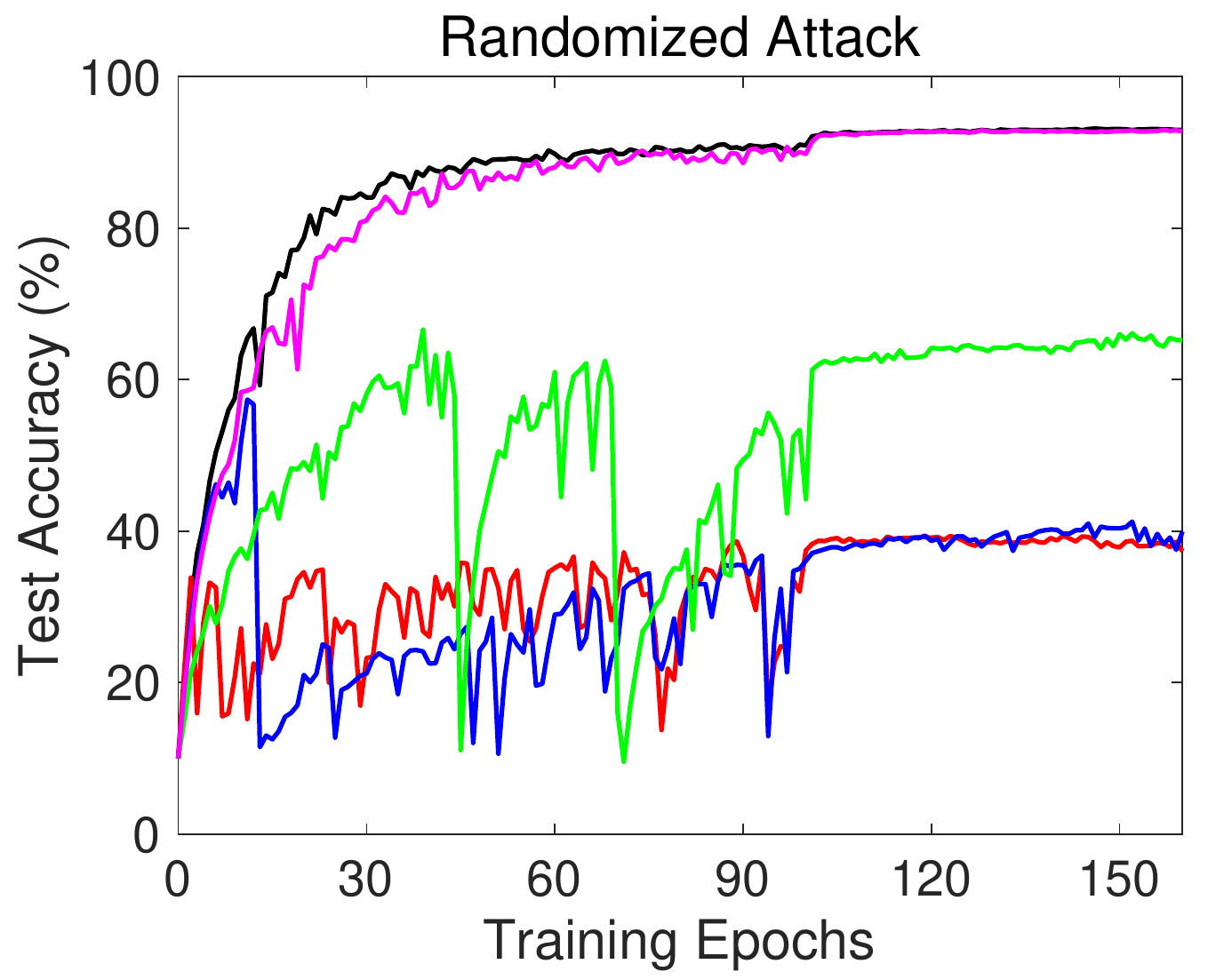}
	}
	%	\quad
	\caption{Defense comparison under time-varying attacks. SignGuard can ensure safe training and achieve decent model accuracy. }
	\label{fig:acc_random_attack}
\end{figure}

\vspace{1ex}
\noindent \textbf{Time-varying attack strategy.} Further, we test different defense algorithms under the time-varying Byzantine attack strategy. We still use the default system setting and change the attack method randomly at each epoch (including no attack scenario). The test accuracy curves of CNN on Fashion-MNIST and ResNet-18 on CIFAR-10 are presented in Fig.~\ref{fig:acc_random_attack}, where the baseline is training under no attack and no defense, and we only test the state-of-the-art defenses. It can be found that our SignGuard could ensure successful model training and closely follow the baseline, while other defenses resulted in significant accuracy fluctuation and model deterioration. For CNN, the training process even collapsed for other defenses, which further demonstrated the superiority of SignGuard.

\begin{figure*}[h]
	\centering
	\subfigure[CNN on Fashion-MNIST]{
		\includegraphics[width=1.9\columnwidth]{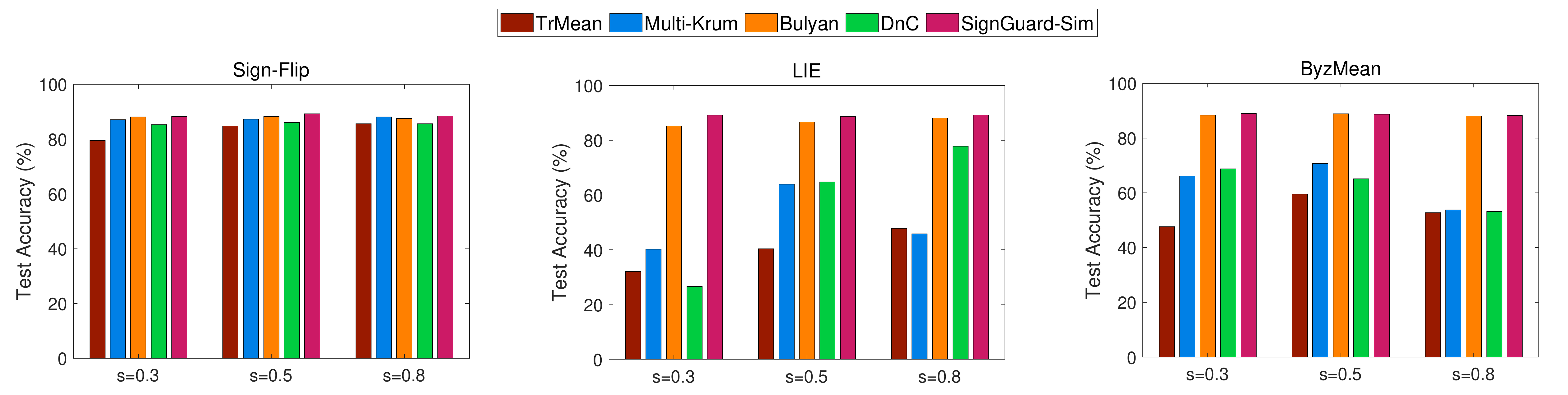}
	}
	%	\quad
	\subfigure[ResNet-18 on CIFAR-10]{
		\includegraphics[width=1.9\columnwidth]{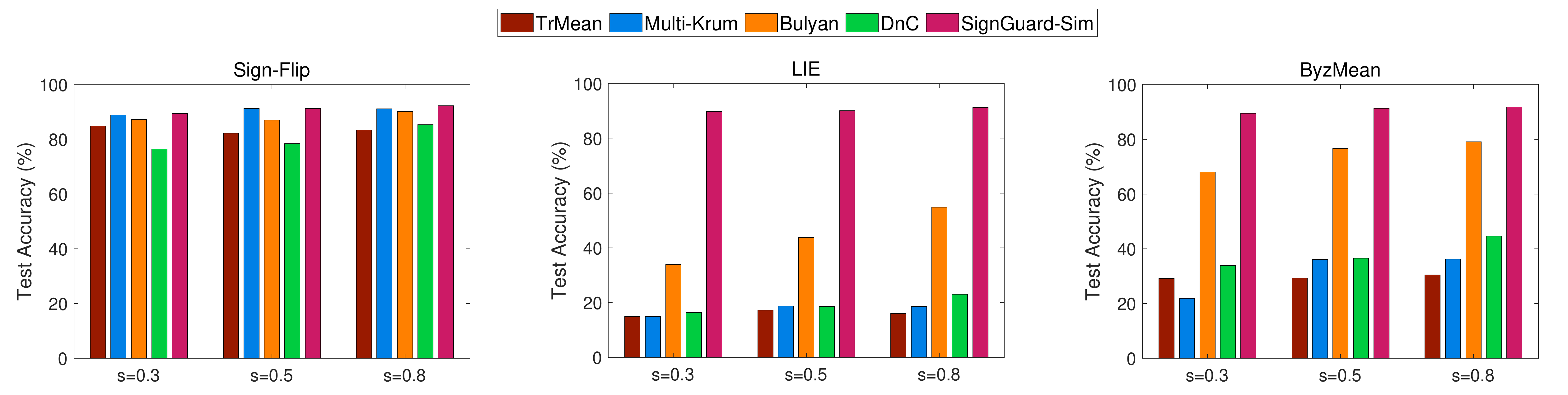}
	}
	\caption{Model accuracy comparison under various attacks and different degrees of non-IID. SignGuard-Sim has the best performance compared with other start-of-the-art defenses. }
	\label{fig:acc_noniid}
	\vspace{-2ex}
\end{figure*}

\subsection{Main Results in Non-IID Settings}

The Byzantine mitigation in non-IID FL settings has been a well-known challenge due to the diversity of gradients. We evaluate our SignGuard-Sim method in synthetic non-IID partitions of Fashion-MNIST and CIFAR-10 datasets. As in previous works, we simulate the non-IID data distribution between clients by allocating $s$-fraction of the dataset in a IID fashion and the remaining (1-$s$)-fraction in a sort-and-partition manner. Specifically, we first randomly select $s$-proportion of the whole training data and evenly distribute them to all clients. Then, we sort the remaining data by labels and divide them into multiple shards, while data in the same shard has the same label, after which each client is randomly allocated with 2 different shards. The parameter $s$ can be used to measure the skewness of data distribution and smaller $s$ will generate more skewed data distribution among clients. We consider three levels of skewness with $s$ = 0.3, 0.5, 0.8, respectively. 

\vspace{1ex}
\noindent \textbf{Efficacy on non-IID data.} We choose the SignGuard-Sim algorithm and compare it with various start-of-the-art defenses. As shown in Fig.~\ref{fig:acc_noniid}, our method still works well under strong attacks in non-IID settings, achieving satisfactory accuracy results in various scenarios. In contrast, TrMean and Multi-Krum could not defend against the LIE attack and ByzMean attack, making them not reliable anymore. Bulyan has a good performance on CNN trained on Fashion-MNIST, but is ineffective under LIE attack on ResNet-18 trained on CIFAR-10. DnC can defend against sign-flipping attack well, but performs poorly on the other scenarios. Those results in non-IID settings further demonstrate the general validness of sign statistics.
\begin{table}[ht] \scriptsize
	\vspace{-2ex}
	\caption{Results under Different Defensive Components}
	\centering
	\label{tab:ablation} 
	
	\newcommand{\tabincell}[2]{\begin{tabular}{@{}#1@{}}#2\end{tabular}} 
	\begin{tabular}{| p{1.3cm}<{\centering} | c | c | p{0.7cm}<{\centering} | p{0.7cm}<{\centering} | p{0.6cm}<{\centering} |}
		\hline
		\multirow{2}{*}{\tabincell{c} {\textbf{Thresholding}}} & 
		\multirow{2}{*}{\tabincell{c} {\textbf{Clustering}}} & 
		\multirow{2}{*}{\tabincell{c} {\textbf{Norm-Clip}}} & 
		\multicolumn{3}{c|}{\textbf{Attacks}}\\
		\cline{4-5} 
		\cline{5-6} 
		& & & Random & Reverse & LIE \\
		\hline
		\Checkmark &  &  & 47.41 & 44.48 & 56.74 \\
		& \Checkmark &  & 88.43 & 25.29 & 88.18 \\
		&  & \Checkmark & 55.27 & 54.29 & 45.98 \\
		\hline
		\Checkmark & \Checkmark &  & \textbf{93.17} & 92.43 & \textbf{93.21} \\
		& \Checkmark & \Checkmark & \underline{93.11} & 93.02 & \underline{93.17} \\
		\hline
		\Checkmark & \Checkmark & \Checkmark & 92.76 & \textbf{93.16} & 92.40 \\
		\hline
	\end{tabular}
	\vspace{-1ex}
\end{table}

\vspace{-1ex}
\subsection{Ablation Study}
Although the above numerical results demonstrate the effectiveness and superiority of our proposed SignGuard framework, our method consists of multiple components and their individual efficacy need more investigation. In this part, we provide some ablation studies under the IID training setting on CIFAR-10 dataset to evaluate the utilities of different defensive components in SignGuard-Sim, including norm-based thresholding, clustering-based filtering, and norm-clipping. Specially, we test the ``Reverse Attack with Scaling"\cite{Rajput19Detox}, in which the Byzantine clients scale the sign-flipped gradient with a positive coefficient $r$, which is selected as the upper bound $R$ of the norm-based thresholding or $r=100$ when no thresholding/norm-clipping is applied. And we also test the random attack and LIE attack. From the results in Table~\ref{tab:ablation}, we could see that every single component could not effectively mitigate all attacks, but clustering-based filtering combined with either thresholding or norm-clipping is capable of defending against a wide range of Byzantine attacks. At first glance, the thresholding and norm-clipping may seem to be redundant since they have similar utilities, however, we believe that the thresholding incurs almost negligible computation cost and could be used to quickly detect those malicious gradients with significantly larger norms. 
\vspace{-1ex}

\section{Conclusion and Future Work}\label{secConclusion}
In this work, we propose a novel Byzantine attack detection framework, namely SignGuard, to mitigate malicious gradients in federated learning systems. It can overcome the drawbacks of the median- and distance-based approaches which are vulnerable to well-crafted attacks and unlike validation-based approaches that require extra data collection in PS. It also does not depend on historical data or other external information, only utilizing magnitude and robust sign statistics from current local gradients, making it a practical way to defend against a variety of model poisoning attacks. Extensive experimental results on image and text classification tasks verify our theoretical and empirical findings, demonstrating the extraordinary effectiveness of our proposed SignGuard-type algorithms. Future directions include developing strategies to defend dynamic and hybrid model poisoning attacks as well as white-box and adaptive attacks in more complex scenarios. And how to design more effective and robust filters in the SignGuard framework for real-world learning systems is also left as an open problem.

\section{Acknowledgment}\label{secAcknowledgment}
The research of Dr. Shao-Lun Huang is supported in part by the Shenzhen Science and Technology Program under Grant KQTD20170810150821146, National Key R\&D Program of China under Grant 2021YFA0715202 and High-end Foreign Expert Talent Introduction Plan under Grant G2021032013L.
The work of Dr. Linqi Song is supported in part by the Hong Kong RGC grant ECS 21212419, InnoHK initiative, the Government of the HKSAR, and Laboratory for AI-Powered Financial Technologies.

\bibliographystyle{IEEEtran}
\bibliography{mybib2021}
\appendix

\subsection{Proof of Proposition 1}\label{appendix:prop1} 

Notice that the standard deviation is estimated on distributed gradients, that is:

\begin{equation*}
	\|std(g^{\{i\in [n]\}})\|^2 = \frac{1}{n}\sum_{i=1}^{n}\|g^{(i)}-\frac{1}{n}\sum_{j=1}^{n}g^{(j)}\|^2
\end{equation*}
so we have:
\begin{equation*}
	\begin{aligned}
		&\mathbb{E}[\|g_m-\tilde{g}\|^2] =\mathbb{E}[\|z\cdot std(g^{\{i\in [n]\}})\|^2]\\
		&=\mathbb{E}[\frac{z^2}{n}\sum_{i=1}^{n}\|g^{(i)}-\frac{1}{n}\sum_{j=1}^{n}g^{(j)}\|^2] \\
		&=\mathbb{E}[\frac{z^2}{n}\sum_{i=1}^{n}\|g^{(i)}-\nabla F(\mathbf{x})+\nabla F(\mathbf{x})-\frac{1}{n}\sum_{j=1}^{n}g^{(j)}\|^2] \\
		&\le\mathbb{E}[\frac{z^2}{n}\sum_{i=1}^{n}\|g^{(i)}-\nabla F(\mathbf{x})\|^2+\|\nabla F(\mathbf{x})-\frac{1}{n}\sum_{j=1}^{n}g^{(j)}\|^2] \\
		&\le\left(1+\frac{1}{n}\right)z^2 \sigma^2
	\end{aligned}
\end{equation*}
%In Appendix~\ref{appendix:prop1}, we already show that:
%\begin{equation*}
%\begin{aligned}
%   \mathbb{E}[\|g_m-\tilde{g}\|^2] \le\left(1+\frac{1}{n}\right)z^2 \sigma^2
%\end{aligned}
%\end{equation*}
and it's easy to see that:
\begin{equation*}
\begin{aligned}
&\mathbb{E}[\|g^{(i)}-\tilde{g}\|^2]=\mathbb{E}[\|g^{(i)}-\frac{1}{n}\sum_{j=1}^{n}g^{(j)}\|^2] \\
&=\mathbb{E}[\|g^{(i)}-\nabla F(\mathbf{x})+\nabla F(\mathbf{x})-\frac{1}{n}\sum_{j=1}^{n}g^{(j)}\|^2] \\
&\le\mathbb{E}[\|g^{(i)}-\nabla F(\mathbf{x})\|^2+\|\nabla F(\mathbf{x})-\frac{1}{n}\sum_{j=1}^{n}g^{(j)}\|^2] \\
&\le\left(1+\frac{1}{n}\right) \sigma^2
\end{aligned}
\end{equation*}
Hence, given a small enough $z$, it's possible for the malicious gradient to have a smaller distance from the true averaged gradient than that of an honest gradient.

Next, we can express the cosine-similarity between malicious gradient and true averaged gradient as well as that of an honest gradient as follows:

\begin{equation*}
\cos(g_m,\tilde{g})=\frac{\left\|g_m\right\|^2+\left\|\tilde{g}\right\|^2-\left\|g_m-\tilde{g}\right\|^2}{2\left\|g_m\right\|\left\|\tilde{g}\right\|}
\end{equation*}

\begin{equation*}
\cos(g^{(i)},\tilde{g})=\frac{\left\|g^{(i)}\right\|^2+\left\|\tilde{g}\right\|^2-\left\|g^{(i)}-\tilde{g}\right\|^2}{2\left\|g^{(i)}\right\|\left\|\tilde{g}\right\|}
\end{equation*}

We can prove that it's possible for the norm of malicious gradient and the norm of certain honest gradient to have following relations:
\begin{equation*}
	\begin{aligned}
	&\left\|g_m\right\| = \xi_m \left\|\tilde{g}\right\| ,~~\|g^{(i)}\| = \xi_i \left\|\tilde{g}\right\|, ~~1\le \xi_i < \xi_m
	\end{aligned}
\end{equation*}
By Jensen inequality, we have:
\begin{equation*}
\left\|\tilde{g}\right\| = \left\|\frac{1}{n}\sum_{i=1}^{n}g^{(i)}\right\| \le \frac{1}{n}\sum_{i=1}^{n} \left\|g^{(i)}\right\| \le \max\{\|g^{(i)}\|\}
\end{equation*}
which means the norm of true averaged gradient is smaller than the averaged norm of honest gradients, so some honest gradients could have bigger norm than $\tilde{g}$, i.e. $\xi_i\ge 1$.

And a appropriate value of $z$ can make $\xi_m > \xi_i$. It's easy to see that:
\begin{equation*}
\begin{aligned}
&\left\|g_m\right\|^2 > \xi_i^2 \left\|\tilde{g}\right\|^2 \\
&\Longleftrightarrow ~~ \sum_{j=1}^{d}(\mu_j-z\sigma_j)^2 > \xi_i^2 \sum_{j=1}^{d}(\mu_j)^2 \\
&\Longleftrightarrow ~~ \sum_{j=1}^{d}(\mu_j^2-2z\mu_j\sigma_j+z^2\sigma_j^2) > \xi_i^2 \sum_{j=1}^{d}(\mu_j)^2 \\
&\Longleftrightarrow ~~ z^2\sum_{j=1}^{d}(\sigma_j^2) -z\sum_{j=1}^{d}(2\mu_j\sigma_j) - (\xi_i^2-1)\sum_{j=1}^{d}(\mu_j)^2 > 0\\
\end{aligned}
\end{equation*}
which obviously holds when given appropriate value of $z$, as the left-hand side is a quadratic function of $z$. 

Therefore, with appropriate selection of $z$, there exists some $i$ such that $1\le \xi_i < \xi_m$. By using these relations of gradient norms, we can get:

\begin{equation*}
	\begin{aligned}
	&\cos(g_m,\tilde{g}) - \cos(g^{(i)},\tilde{g})\\
	&= \frac{(\xi_m^2+1)\left\|\tilde{g}\right\|^2-\left\|g_m-\tilde{g}\right\|^2}{2\xi_m\left\|\tilde{g}\right\|^2} - \frac{(\xi_i^2+1)\left\|\tilde{g}\right\|^2-\left\|g^{(i)}-\tilde{g}\right\|^2}{2\xi_i\left\|\tilde{g}\right\|^2}\\
	&> \left(\frac{(\xi_m^2+1)}{2\xi_m}-\frac{(\xi_i^2+1)}{2\xi_i}\right)+\frac{\left\|g_m-\tilde{g}\right\|^2}{2\left\|\tilde{g}\right\|^2}\left(\frac{1}{\xi_i}-\frac{1}{\xi_m}\right)\\
	&=\frac{(\xi_m-\xi_i)(\xi_m \xi_i -1)}{2\xi_m \xi_i}+\frac{(\xi_m-\xi_i)\left\|g_m-\tilde{g}\right\|^2}{2\xi_m \xi_i\left\|\tilde{g}\right\|^2}\\
	&>0
	\end{aligned}
\end{equation*}
Hence, it's possible for the malicious gradient to have a bigger cosine-similarity with true averaged gradient than that of an honest gradient.

\subsection{Proof of Lemma 1}
\label{appendix:lemm1} 
Given a arbitrary subset of clients $\mathcal{G}$ with $|\mathcal{G}|=(1-\beta)n$ and $\beta<0.5$.
Let $\mathbf{A}=\sum\limits_{i\notin \mathcal{G}}\left(g_{t}^{(i)}-\nabla F(\mathbf{x}_{t})\right)$, $\mathbf{B}=\sum\limits_{j\in \mathcal{G}}\left(g_{t}^{(j)}-\nabla F(\mathbf{x}_{t})\right)$, then $\mathbf{A}$ and $\mathbf{B}$ are independent. We have $\mathbb{E}[\mathbf{A}+\mathbf{B}]=\mathbf{0}$.
Recall that $\sigma^2$ is the bounded local variance for local gradient and $\kappa^2$ is bounded deviation between local and global gradient. Applying the Jensen inequality, we have
\begin{equation*}
\begin{aligned}
\left\|\mathbb{E}\left[\mathbf{A}\right]\right\|^2 &\leq \beta n\sum\limits_{i\notin \mathcal{G}}\left\|\nabla F_i(\mathbf{x}_{t})-\nabla F(\mathbf{x}_{t})\right\|^2 \leq \beta^2n^2\kappa^2\\
\left\|\mathbb{E}\left[\mathbf{B}\right]\right\|^2 &\leq (1-\beta)n\sum\limits_{i\in \mathcal{G}}\left\|\nabla F_i(\mathbf{x}_{t})-\nabla F(\mathbf{x}_{t})\right\|^2 \leq (1-\beta)^2n^2\kappa^2\\
\end{aligned}
\end{equation*}
Notice that $\mathbb{E}[\mathbf{A}]=-\mathbb{E}[\mathbf{B}]$, thus
\begin{equation*}
\left\|\mathbb{E}\left[\mathbf{A}\right]\right\|^2=\left\|\mathbb{E}\left[\mathbf{B}\right]\right\|^2\leq \min\{\beta^2 n^2\kappa^2, (1-\beta)^2n^2\kappa^2\} = \beta^2 n^2\kappa^2
\end{equation*}
Using the basic relation between expectation and variance, we have
\begin{equation*}
\begin{aligned}
&\mathbb{E}\left\|\mathbf{A}\right\|^2=\left\|\mathbb{E}[\mathbf{A}]\right\|^2+\text{var}[\mathbf{A}]\leq\left\|\mathbb{E}[\mathbf{A}]\right\|^2+\beta n\sigma^2\\
&\mathbb{E}\left\|\mathbf{B}\right\|^2=\left\|\mathbb{E}[\mathbf{B}]\right\|^2+\text{var}[\mathbf{B}]\leq\left\|\mathbb{E}[\mathbf{B}]\right\|^2+(1-\beta) n\sigma^2
\end{aligned}
\end{equation*}
which leads to
\begin{equation*}
\begin{aligned}
\mathbb{E}\left\|\mathbf{B}\right\|^2 \le \beta^2 n^2\kappa^2 +(1-\beta)n\sigma^2
\end{aligned}
\end{equation*}
Then, we directly have
\begin{equation*}
\begin{aligned}
&\mathbb{E}\left[\left\|\frac{1}{|\mathcal{G}|}\sum\limits_{i\in \mathcal{G}}\left(g_{t}^{(i)}\right)-\nabla F(\mathbf{x}_{t})\right\|^2\right]=\frac{1}{(1-\beta)^2n^2}\mathbb{E}\left\|\mathbf{B}\right\|^2\\
&\leq { \frac{\beta^2\kappa^2}{(1-\beta)^2}}+\frac{\sigma^2}{(1-\beta)n}
\end{aligned}
\end{equation*}
It completes the proof of Lemma 1.

%\vspace{2ex}
\subsection{Proof of Theorem 1} 
\label{append:theorem1}

Taking the total expectations of averaged gradient on local sampling and randomness in aggregation rule, we have
\begin{equation*}
\begin{aligned}
&\mathbb{E}_t[F(\mathbf{x}_{t+1})] - F(\mathbf{x}_t) \\
&\leq -\eta \left\langle \nabla F(\mathbf{x}_t),\mathbb{E}_t\left[\hat{g}_t\right]\right\rangle+\frac{L\eta^2}{2}\mathbb{E}_t\left[\left\|\hat{g}_t\right\|^2\right]\\
&= -\eta \left\langle \nabla F(\mathbf{x}_t),\mathbb{E}_t\left[\hat{g}_t-\tilde{g}_t+\tilde{g}_t-\nabla F(\mathbf{x}_t)+\nabla F(\mathbf{x}_t)\right]\right\rangle\\
&\qquad+\frac{L\eta^2}{2}\mathbb{E}_t\left[\left\|\hat{g}_t-\nabla F(\mathbf{x}_t)+\nabla F(\mathbf{x}_t)\right\|^2\right]\\
&\leq -\eta \left\langle \nabla F(\mathbf{x}_t),\mathbb{E}_t\left[\hat{g}_t-\tilde{g}_t\right]\right\rangle -\eta \left\langle \nabla F(\mathbf{x}_t),\mathbb{E}_t\left[\tilde{g}_t-\nabla F(\mathbf{x}_t)\right]\right\rangle\\
&\qquad-\eta\left\|\nabla F(\mathbf{x}_t)\right\|^2 +L\eta^2\left\|\nabla F(\mathbf{x}_t)\right\|^2+L\eta^2\mathbb{E}_t\left[\left\|\hat{g}_t-\nabla F(\mathbf{x}_t)\right\|^2\right]\\
\end{aligned}
\end{equation*}
From Assumption 1 \& 2, we have
\begin{equation*}
\begin{aligned}
\left[\mathbb{E}\left\|\hat{g}_t-\bar{g}_t\right\|\right]^2
\leq c{\delta}\sup_{i,j\in \mathcal{G}}\mathbb{E}[\|{g_t^{(i)}}-{g_t^{(j)}}\|^2]\leq 2c{\delta}(\sigma^2+\kappa^2)
\end{aligned}
\end{equation*}
then by Young's Inequality with $\rho=2$, we can get
\begin{equation*}
\begin{aligned}
&-\eta \left\langle \nabla F(\mathbf{x}_t),\mathbb{E}_t\left[\hat{g}_t-\tilde{g}_t\right]\right\rangle\\
&\leq \eta \left\|\nabla F(\mathbf{x}_t)\right\| \cdot \mathbb{E}_t\left\|\hat{g}_t-\tilde{g}_t\right\|\\
& \leq\frac{\sqrt{\delta}\eta}{2\rho}\left\|\nabla F(\mathbf{x}_t)\right\|^2 + \frac{\rho}{2}\cdot 2\sqrt{\delta}\eta c(\sigma^2+\kappa^2)\\
&\leq\frac{\sqrt{\delta}\eta}{4}\left\|\nabla F(\mathbf{x}_t)\right\|^2 + 2\sqrt{\delta}\eta c(\sigma^2+\kappa^2)
\end{aligned}
\end{equation*}
Combining with Lemma 2, we get
\begin{equation*}
\begin{aligned}
&-\eta \left\langle \nabla F(\mathbf{x}_t),\mathbb{E}_t\left[\tilde{g}_t-\nabla F(\mathbf{x}_t)\right]\right\rangle\\
&\leq \eta \left\|\nabla F(\mathbf{x}_t)\right\| \cdot \mathbb{E}_t\left\|\tilde{g}_t-\nabla F(\mathbf{x}_t)\right\|\\
&\leq\frac{\beta\eta}{2}\left\|\nabla F(\mathbf{x}_t)\right\|^2 + \frac{\beta\eta\kappa^2}{2(1-\beta)^2}
\end{aligned}
\end{equation*}
and
\begin{equation*}
\begin{aligned}
&\mathbb{E}_t\left[\left\|\hat{g}_t-\nabla F(\mathbf{x}_t)\right\|^2\right] \\
&= \mathbb{E}_t\left[\left\|\hat{g}_t-\bar{g}_t + \bar{g}_t - \nabla F(\mathbf{x}_t)\right\|^2\right]\\
&\leq 2\mathbb{E}_t\left[\left\|\hat{g}_t-\bar{g}_t\right\|^2\right]+2\mathbb{E}_t\left[\left\|\bar{g}_t - \nabla F(\mathbf{x}_t)\right\|^2\right]\\
&=2\left[\mathbb{E}\left\|\hat{g}_t-\bar{g}_t\right\|\right]^2+2\text{var}\left\|\hat{g}_t\right\|+2\mathbb{E}_t\left[\left\|\bar{g}_t - \nabla F(\mathbf{x}_t)\right\|^2\right]\\
&\leq \quad \begin{matrix} \underbrace{ 4c\delta(\sigma^2+\kappa^2)+2b^2+\frac{2\beta^2\kappa^2}{(1-\beta)^2}+\frac{2\sigma^2}{(1-\beta)n} } \\ =\Delta_1 \end{matrix}
\end{aligned}
\end{equation*}
In the above derivations, the basic inequality $2\mathbf{a}\cdot \mathbf{b}\leq \mathbf{a}^2+\mathbf{b}^2$ is applied. Taking total expectation and rearranging the terms, we get
\begin{equation*}
\begin{aligned}
&\eta\left(\frac{4-\sqrt{\delta}-2\beta}{4}-L\eta \right)\mathbb{E}[\left\|\nabla F(\mathbf{x}_t)\right\|^2]\leq \mathbb{E}[F(\mathbf{x}_{t})-F(\mathbf{x}_{t+1})]\\
&\qquad\qquad+2\sqrt{\delta}\eta c(\sigma^2+\kappa^2)+\frac{\beta\eta\kappa^2}{2(1-\beta)^2}+L\eta^2\Delta_1\\
\end{aligned}
\end{equation*}
Assume that $\eta \le (2-\sqrt{\delta}-2\beta)/(4L)$, thus we have $\left(\frac{4-\sqrt{\delta}-2\beta}{4}-L\eta \right)\geq\dfrac{1}{2}$. Taking summation and dividing by $\eta\left(\frac{4-\sqrt{\delta}-2\beta}{4}-L\eta \right)T$, then we finally get
\begin{equation*}
\begin{aligned}
&\dfrac{1}{T}\sum_{t=0}^{T-1}\mathbb{E}[\left\|\nabla F(\mathbf{x}_t)\right\|^2]\leq \frac{2(F(\mathbf{x}_0)-F^*)}{\eta T}+2L\eta\Delta_1\\
&\qquad\qquad\qquad\qquad+\quad
\begin{matrix} \underbrace{ 4\sqrt{\delta} c(\sigma^2+\kappa^2)+\frac{\beta\kappa^2}{(1-\beta)^2} } \\ =\Delta_2 \end{matrix}
\end{aligned}
\end{equation*}
which completes the proof.

\end {document}